\documentclass[aip,cha,reprint,amsmath,amssymb,nofootinbib,floatfix]{revtex4-2}

\usepackage{graphicx}%
\usepackage{multirow}%
\usepackage{amsthm}%
\usepackage{xcolor}%
\usepackage{booktabs}%
\usepackage{bm}%

\theoremstyle{plain}%
\newtheorem{proposition}{Proposition}%
\newtheorem{corollary}{Corollary}%
\theoremstyle{remark}%
\newtheorem{remark}{Remark}%

\newcommand{\dhip}{d_h}
\newcommand{\rv}{r^{v}}
\newcommand{\thv}{\theta^{v}}
\newcommand{\rvz}{r^{v}_0}
\newcommand{\td}{\mathrm{td}}
\newcommand{\lo}{\mathrm{lo}}
\newcommand{\X}{\mathbf{X}}
\newcommand{\PE}{\mathrm{PE}}
\newcommand{\Ntd}{N_{\td}}

\begin{document}

\title[Central pivot point control of a body-attached spring-mass runner]{Central pivot point control of a body-attached spring-mass runner: periodic gaits, stabilization, template embedding and approximate analytic return maps}

\author{{\.I}zel Sever}
\email{izel.sever@gmail.com}
\affiliation{Department of Electrical and Electronics Engineering, Middle East Technical University, 06800 Ankara, T\"urkiye}
\affiliation{Roketsan Inc., Ankara, T\"urkiye}

\author{O. Kaan Karag\"oz}
\email{karagoz.osman@metu.edu.tr}
\affiliation{Department of Electrical and Electronics Engineering, Middle East Technical University, 06800 Ankara, T\"urkiye}
\affiliation{Center for Robotics and Artificial Intelligence (ROMER), Middle East Technical University, 06800 Ankara, T\"urkiye}

\author{M. Mert Ankaral{\i}}
\email{Author to whom correspondence should be addressed: mertan@metu.edu.tr}
\affiliation{Department of Electrical and Electronics Engineering, Middle East Technical University, 06800 Ankara, T\"urkiye}
\affiliation{Center for Robotics and Artificial Intelligence (ROMER), Middle East Technical University, 06800 Ankara, T\"urkiye}

\date{\today}

\begin{abstract}
The spring-loaded inverted pendulum (SLIP) describes the center-of-mass (COM) motion of runners but has no trunk. In trunk models, the hip torque couples posture and COM motion and destroys the central-force structure on which SLIP analysis relies. As a result, trunk models have been studied almost only numerically. We show that this structure can be restored. The central pivot point (CPP), a hip-torque law that directs the ground reaction force through the COM, decouples pitch exactly and reduces stance to a planar central-force problem. At upright posture, time-reversal symmetry produces one-parameter families of periodic gaits. A stance-phase pitch law and a stride-level regulator stabilize all computed gaits, also with bounded hip torque and friction. Under CPP, any radial force law in virtual-leg coordinates can be embedded independently of posture. For a linear radial template, the stance radius and the constant-momentum sweep have exact closed forms. With a new gravity correction, the resulting analytic return map, to our knowledge the first for a trunk model, predicts liftoff velocity and apex position with mean errors of 1.6\% and 0.9\%. The natural gaits of the embedded runner are symmetric and need no force step at touchdown. A near-reversibility argument explains their unit multiplier, and a scalar equation predicts them within 0.8\%. A deadbeat stride controller based on the analytic map tracks changes of speed, height and posture. CPP thus makes trunk models accessible to the analytic tools developed for SLIP.
\end{abstract}


\maketitle

\begin{quotation}
Running animals and legged robots carry a heavy trunk on a springy leg. Simple spring-mass models capture the bouncing motion of the body well, and their analysis has guided robot design for decades. A trunk breaks this picture: the hip torque that balances the trunk also pushes the body, and the resulting dynamics have been studied almost only by simulation. We show that a simple hip-torque rule, which makes the ground force always point through the center of mass, removes this coupling exactly. The trunk can then be held at any posture while the body bounces like a point mass on a spring. This restores the symmetry and structure of the classical model, gives closed-form predictions of each stride, and explains why the periodic gaits of such runners are only neutrally stable. The results bring the analytic tools of spring-mass running to models with a trunk.
\end{quotation}

%% ===================================================================
\section{Introduction}\label{sec:intro}
%% ===================================================================

Running animals and legged robots share complex, hybrid and strongly nonlinear dynamics \cite{holmes2006dynamics}. Full and Koditschek proposed to organize this complexity hierarchically \cite{full1999templates}. Simple \emph{templates} describe a target behavior, whereas more detailed \emph{anchors} represent the physical system. A controller that makes the anchor behave like the template allows design and analysis at the template level \cite{raibert1986legged,poulakakis2009spring,ankarali2011control,lu2018clock}.

The spring-loaded inverted pendulum (SLIP) is the most widely used running template \cite{Blickhan1989,Mcmahon1990,holmes2006dynamics}. It consists of a point mass on a massless spring leg. SLIP describes the COM motion of many runners well, and it serves as a control target for robots \cite{raibert1986legged,SaranliRHEX,martin2017experimental}. However, SLIP has no trunk. In humans, humanoids and many robots the hip does not coincide with the COM. Leg forces then create a pitching moment that must be balanced. Existing remedies include reactive pitch controllers \cite{ankarali2011control}, passive mechanisms such as bisecting hips and bow legs \cite{wisse2007adding,brown1998bow}, and posture laws based on a virtual pivot point (VPP) \cite{maus2010upright,sharbafi2013robust,drama2020trunk,drama2020postural}. When such posture control is active, the COM generally no longer follows SLIP trajectories \cite{martin2017experimental}.

Templates that include the trunk address this issue. Poulakakis and Grizzle introduced the asymmetric SLIP (ASLIP), a SLIP-like model with a rigid trunk and a hip offset \cite{poulakakis2007formal,poulakakis2009spring}. They used hip torque and leg force to create an invariant manifold on which the COM obeys SLIP dynamics. The same morphology is called trunk-SLIP (TSLIP) \cite{sharbafi2013robust}; we use this name below. Maus et al.\ showed that ground reaction forces (GRFs) of humans and animals in walking and running point to a VPP above the COM \cite{maus2010upright}. A hip torque that reproduces this pattern stabilizes the trunk like a pendulum \cite{maus2010upright,sharbafi2013robust}. Averaging arguments confirm this stabilization, and VPP posture control has been transferred to a quadruped robot \cite{sun2023posture}. We previously analyzed the periodic gaits of TSLIP with a VPP above the COM and designed a stride-level controller for them \cite{karagoz2022vpp}. More recent measurements show that the picture is not uniform. In human running, the GRFs intersect \emph{below} the COM in the sagittal plane \cite{drama2020postural,firouzi2022human}. The location of this point redistributes work between hip and leg and shapes trunk pitch oscillations \cite{drama2020trunk}. Simulations also suggest that a GRF intersection point is not necessary for stable walking \cite{schreff2023vpp}. Trunk templates continue to be extended, for example with foot mass and leg retraction \cite{ossadnik2024ult}. Their analysis, however, remains almost entirely numerical.

Analytic approximations play a central role in SLIP research. The SLIP stance dynamics are not integrable, so return maps are usually computed by numerical integration. Approximate closed-form stance maps reveal how parameters shape the gait, and they support fast planning and control \cite{Schwind2000,GEYER2005315,ghigliazza2005simply,Arslan2009,Saranli2010,Yu2012,Yu2021,Shahbazi2016,arslan2012reactive}. Gravity corrections extend these approximations to asymmetric strides \cite{Arslan2009,Saranli2010}. Recent work has extended them to three dimensions \cite{yu2019seeking}, to articulated legs \cite{chen2023articulated} and to large asymmetric steps \cite{kilic2023novel}. Asymptotic methods give the stiffness of periodic gaits \cite{plociniczak2020asymptotic} and conditions for their stability \cite{wroblewska2021stability}. Selvitella and Foster assumed a constant angular velocity in stance, which makes the radial motion elementary and yields explicit stability conditions for symmetric gaits \cite{selvitella2024approximate}. Physics-guided neural networks offer a data-driven alternative \cite{ozturk2026physics}. Partial feedback linearization (PFL) of the radial dynamics yields exact or simpler solutions when the leg is actuated \cite{Piovan2016,Hamzacebi2017}. These results, however, are limited to point-mass SLIP models without a trunk. To our knowledge, no analytic approximations of the stance map exist for trunk models. The reason is structural: trunk coupling destroys the central-force structure on which all SLIP approximations rely.

In this paper, we study a hip-torque policy that restores this structure. The policy, which we call the central pivot point (CPP), directs the GRF through the COM during the whole stance phase \cite{karagoz2020aim}. CPP can be viewed as the limiting case of the VPP concept in which the pivot coincides with the COM \cite{maus2010upright}. It lies between the pivot above the COM of classical VPP models and the intersection point below the COM reported for human running \cite{drama2020postural,firouzi2022human}. Its value lies less in posture stabilization than in the exact decoupling it creates. This decoupling allows a structural analysis of the gait, a posture-independent template embedding, and analytic approximations of the stance map. In other words, CPP brings the analytic toolbox of SLIP to a model with a trunk.

This paper is based on the MSc thesis of the first author \cite{sever2022thesis}. The thesis developed the core of this work: the CPP policy, the hierarchical gait controller, the virtual-leg embedding with a linear radial template, approximate analytic return maps with gravity corrections, and the simulations and analyses that test them. A first part, covering the CPP concept, the open-loop fixed points and the PD--LQR controller, appeared in \cite{karagoz2020aim}. A VPP above the COM was treated in \cite{karagoz2020thesis,karagoz2022vpp}. Here we present the complete work, recompute all results, and extend it in the following directions.

On the CPP side, we prove that CPP decouples pitch exactly and reduces stance to a planar central-force problem (Proposition~\ref{prop:decoupling}). We characterize the structure of the apex return map: pitch rate is invariant, pitch drifts, and the map has structural unit multipliers. At upright posture, time-reversal symmetry generates one-parameter families of periodic gaits (Propositions~\ref{prop:structure} and~\ref{prop:symmetry}). This reduces the gait search to a scalar equation and leads to a corrected stability characterization based on the transverse multiplier. We also evaluate the PD--LQR controller under hip-torque and friction limits.

On the embedding side, we show that CPP allows any radial force law in virtual-leg coordinates to be embedded independently of posture (Proposition~\ref{prop:embedding}). We compare a linear and a SLIP template and state when the linear template is physically admissible. For the linear template, we integrate the constant-momentum sweep exactly and introduce a new gravity correction. We validate the resulting map on more than 13{,}000 admissible strides and compare it with earlier corrections and with a SLIP-template approximation. We then explain the natural gaits of the embedded runner by a near-reversibility argument (Proposition~\ref{prop:natural}) and predict them from a scalar equation. Finally, we use the analytic map for deadbeat stride control with a filtered error feedback.

These results matter beyond the specific model. First, CPP separates two tasks that are usually coupled: the leg shapes the COM motion, and the hip holds the trunk at any desired posture. A robot can therefore lean forward to accelerate or carry a load without changing its stride dynamics. Second, the closed-form stride map costs a few microseconds. It is fast enough for online planning, deadbeat control and model-predictive control, which so far relied on SLIP models without a trunk. Third, the structural results are not tied to CPP. Reversibility and near-reversibility explain why non-hyperbolic gaits and gait families appear in legged models that are neither passive nor conservative. This view can guide the analysis of other actuated hybrid systems.

Section~\ref{sec:model} introduces the model and the numerical methods. Section~\ref{sec:cpp} presents CPP and its structural properties, and Section~\ref{sec:control} the gait controller. Section~\ref{sec:embedding} introduces the template embedding, and Section~\ref{sec:approx} derives the approximate return map. Section~\ref{sec:prediction} reports its prediction performance. Section~\ref{sec:natural} studies the natural gaits of the embedded runner, and Section~\ref{sec:mapcontrol} the map-based controller. Section~\ref{sec:discussion} discusses the results and their limitations.

%% ===================================================================
\section{Trunk-SLIP model}\label{sec:model}
%% ===================================================================

\begin{figure*}[!t]
\centering
\includegraphics[width=\textwidth]{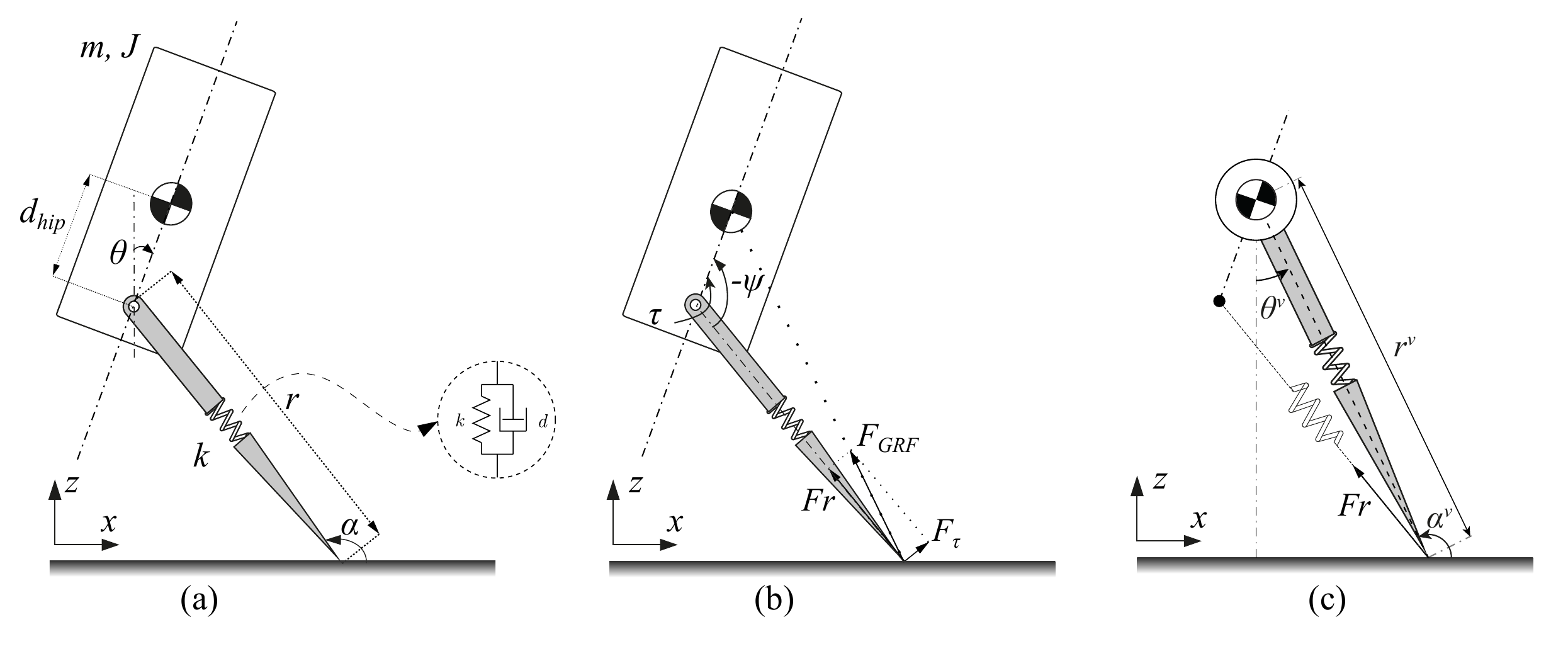}
\caption{(a) Trunk-SLIP (TSLIP) model and its parameters; the labels $d_{hip}$ and $d$ correspond to $\dhip$ and $c$ in the text. (b) Central pivot point (CPP): the hip torque $\tau$ adds a force component $F_\tau=\tau/r$ normal to the leg, so that the GRF passes through the COM. (c) Virtual leg from the toe to the COM, with length $\rv$ and angle $\thv$}
\label{fig:model}
\end{figure*}

The TSLIP model is shown in Fig.~\ref{fig:model}(a). A rigid trunk of mass $m$ and pitch inertia $J$ about its COM carries a massless leg. The hip lies at distance $\dhip$ below the COM along the trunk axis. The leg has a linear spring of stiffness $k$ and rest length $r_0$, and an optional radial damper with coefficient $c$. A torque $\tau$ acts between trunk and leg at the hip. We use the generalized coordinates $\mathbf{q}=[x,\,z,\,\theta]^T$, where $(x,z)$ is the COM position and $\theta$ is the trunk pitch, positive for forward lean. Table~\ref{tab:params} lists the parameter values, which correspond to a human-sized runner \cite{sharbafi2013robust}.

\begin{table}[!tb]
\caption{Physical parameters of the TSLIP model}\label{tab:params}
\begin{tabular}{@{}llr@{}}
\toprule
Parameter & Symbol & Value\\
\midrule
Body mass & $m$ & 80\,kg\\
Pitch inertia about the COM & $J$ & 5\,kg\,m$^2$\\
COM--hip distance & $\dhip$ & 0.1\,m\\
Leg rest length & $r_0$ & 1\,m\\
Leg stiffness & $k$ & 20\,kN/m\\
Gravitational acceleration & $g$ & 9.8\,m/s$^2$\\
\botrule
\end{tabular}
\end{table}

\subsection{Phases and equations of motion}

\begin{figure}[!tb]
\centering
\includegraphics[width=\columnwidth]{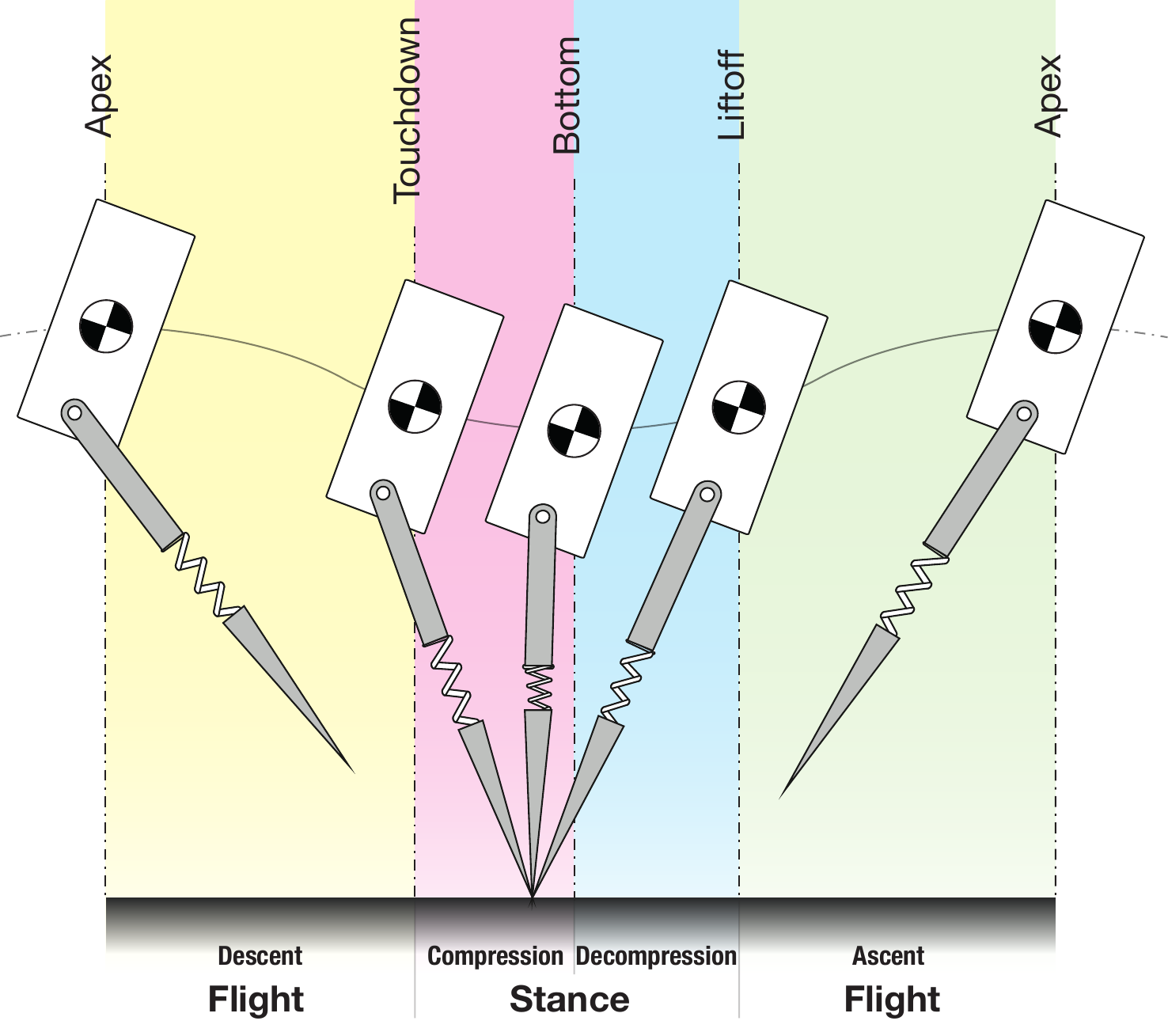}
\caption{Phases and events of a TSLIP stride: apex, touchdown, bottom, liftoff and next apex}
\label{fig:phases}
\end{figure}

The model alternates between flight and stance (Fig.~\ref{fig:phases}). In flight, the COM follows a ballistic path and the pitch rate is constant:
\begin{equation}
\ddot x=0,\qquad \ddot z=-g,\qquad \ddot\theta=0.
\label{eq:flight}
\end{equation}
The massless leg exerts no torque on the trunk in flight. It is placed at the commanded touchdown angle $\alpha_\td$ before touchdown.

In stance, the toe is a frictionless pin at the origin. The hip position is
\begin{equation}
x_h=x-\dhip\sin\theta,\qquad z_h=z-\dhip\cos\theta,
\label{eq:hip}
\end{equation}
and the leg length, leg angle and hip angle are
\begin{equation}
\begin{gathered}
r=\sqrt{x_h^2+z_h^2},\qquad \alpha=\operatorname{atan2}(z_h,x_h),\\ \psi=\alpha+\theta+\tfrac{\pi}{2}.
\end{gathered}
\label{eq:leg}
\end{equation}
The angle $\alpha$ is measured from the ground, so $\alpha>90^\circ$ means that the toe is ahead of the hip. The stance dynamics follow from the Euler--Lagrange equations,
\begin{equation}
\begin{gathered}
\mathbf{M}\ddot{\mathbf{q}}=-mg\,\mathbf{e}_2+\nabla_{\mathbf{q}} r\,F_r+\nabla_{\mathbf{q}}\psi\,\tau,\\ \mathbf{M}=\operatorname{diag}(m,m,J),
\end{gathered}
\label{eq:stance}
\end{equation}
where $\mathbf{e}_2=[0,1,0]^T$ and the gradients are given in Appendix~\ref{app:eom}. The radial leg force is
\begin{equation}
F_r=-k(r-r_0)-c\,\dot r+u_r,
\label{eq:Fr}
\end{equation}
where $u_r$ is an optional actuator force acting along the leg, in parallel with the spring. We use $u_r$ only from Section~\ref{sec:embedding} on. As for SLIP, the stance dynamics \eqref{eq:stance} are not integrable. The GRF acts at the toe and equals
\begin{equation}
\mathbf{F}=\frac{1}{r}\begin{bmatrix}F_r x_h-\tau z_h/r\\ F_r z_h+\tau x_h/r\end{bmatrix}.
\label{eq:grf}
\end{equation}

\subsection{Events and apex return map}

Touchdown occurs when the toe reaches the ground, $z_h-r_0\sin\alpha_\td=0$ with $\dot z<0$. Liftoff occurs when the vertical GRF crosses zero from above. The apex is the flight state with $\dot z=0$ and $\ddot z<0$. We use the apex as Poincar\'e section. The dynamics do not depend on the horizontal position, so the apex state is
\begin{equation}
\X=[z_a,\;\dot x_a,\;\theta_a,\;\dot\theta_a]^T.
\end{equation}
The apex return map is $\X_{k+1}=P(\X_k,\mathbf{u}_k)$, where $\mathbf{u}_k$ collects the stride-level inputs, such as $\alpha_\td$ and $c$. A fixed point $\X^*=P(\X^*,\mathbf{u}^*)$ corresponds to a periodic gait. Its local stability is determined by the eigenvalues (Floquet multipliers) of $\mathbf{A}=\partial P/\partial\X$ at $\X^*$ \cite{holmes2006dynamics}. Flight is solved analytically. Stance is integrated with a fourth-order Runge--Kutta scheme with a step of $10^{-4}$\,s, events are located by bisection, and Jacobians are computed by central differences.

%% ===================================================================
\section{Central pivot point}\label{sec:cpp}
%% ===================================================================

\subsection{Hip torque law, decoupling and energy}

The CPP policy chooses the hip torque so that the GRF passes through the COM. Let $\gamma$ be the signed angle at the toe between the physical leg and the virtual leg that connects the toe to the COM [Fig.~\ref{fig:model}(c)]. The hip torque adds a force $\tau/r$ normal to the leg. The resultant GRF is aligned with the virtual leg if $\tau/r=F_r\tan\gamma$. This gives
\begin{equation}
\tau_c=F_r\,r\tan\gamma,\qquad \tan\gamma=\frac{\dhip\sin\psi}{r-\dhip\cos\psi}.
\label{eq:tau_cpp}
\end{equation}
The derivation of $\gamma$ is given in Appendix~\ref{app:geometry}. The CPP torque is a static function of the leg force and the configuration. It requires no model of the trunk inertia.

We describe the COM by the virtual-leg polar coordinates
\begin{equation}
\rv=\sqrt{x^2+z^2},\qquad \thv=\operatorname{atan2}(-x,z),
\end{equation}
so that $x=-\rv\sin\thv$ and $z=\rv\cos\thv$. The angle $\thv$ is positive when the COM is behind the toe. Let $\Phi=\theta+\thv$ denote the angle between the trunk axis and the virtual leg at the COM. The law of cosines in the triangle toe--hip--COM gives
\begin{equation}
\begin{gathered}
r^2=(\rv)^2+\dhip^2-2\rv\dhip\cos\Phi,\\ r\cos\gamma=\rv-\dhip\cos\Phi.
\end{gathered}
\label{eq:triangle}
\end{equation}

\begin{proposition}[Decoupling]\label{prop:decoupling}
Assume $r>\dhip$ in stance and apply $\tau=\tau_c$ from \eqref{eq:tau_cpp}. Then:
\begin{enumerate}
\item[(i)] the pitch dynamics reduce to $\ddot\theta=0$;
\item[(ii)] the GRF is collinear with the virtual leg and has the magnitude
\begin{equation}
N=\frac{F_r}{\cos\gamma}=\frac{F_r\,r}{\rv-\dhip\cos\Phi};
\label{eq:N}
\end{equation}
\item[(iii)] the COM obeys the planar central-force equations
\begin{align}
m\big(\ddot r^{v}-\rv(\dot\theta^{v})^2+g\cos\thv\big)&=N,\label{eq:cpp_radial}\\
\frac{d}{dt}\big((\rv)^2\dot\theta^{v}\big)&=g\,\rv\sin\thv,\label{eq:cpp_angular}
\end{align}
in which the trunk enters only through $\Phi=\theta+\thv$ in $N$.
\end{enumerate}
\end{proposition}

\begin{proof}
The GRF acts at the toe and is the only contact force. By construction of \eqref{eq:tau_cpp}, it lies on the line from the toe to the COM, so its moment about the COM vanishes. The hip torque is internal to the trunk--leg system and the leg is massless. Hence the net moment on the trunk about its COM is zero, which gives (i). The GRF components along and normal to the leg are $F_r$ and $F_r\tan\gamma$, which gives \eqref{eq:N} with \eqref{eq:triangle}. Newton's law for the COM in polar coordinates gives \eqref{eq:cpp_radial}. The GRF exerts no moment about the toe, so the angular momentum about the toe changes only through gravity, which gives \eqref{eq:cpp_angular}.
\end{proof}

We verified \eqref{eq:tau_cpp}--\eqref{eq:cpp_angular} against the full model \eqref{eq:stance} to machine precision. Proposition~\ref{prop:decoupling} has two consequences. First, the stance phase has the same central-force structure as SLIP, with a configuration-dependent radial force $N$. Second, the pitch becomes a pure double integrator. It can be controlled independently by an additional torque, which is the basis of Section~\ref{sec:control}. Compared with a VPP above the COM, which turns the trunk into a stable pendulum \cite{maus2010upright,sharbafi2013robust}, CPP makes the pitch dynamics neutral.

Let $E=\tfrac12 m(\dot x^2+\dot z^2)+\tfrac12 J\dot\theta^2+mgz+\tfrac12 k(r-r_0)^2$ be the total mechanical energy. Along stance trajectories,
\begin{equation}
\dot E=\tau\dot\psi-c\,\dot r^2+u_r\dot r.
\label{eq:energy}
\end{equation}
Under CPP with $c=0$ and $u_r=0$, the hip actuator is the only source or sink of energy. The work per stance is $W=\int\tau_c\dot\psi\,dt$. The force $N$ in \eqref{eq:N} depends on $\Phi$ and not only on $\rv$, so the COM moves in a non-conservative central force field. The field is symmetric about $\Phi=0$, that is, about the instant at which the hip lies on the virtual leg.

\subsection{Structure of the CPP return map}\label{sec:structure}

We first consider the passive CPP runner, with fixed $\alpha_\td$, $c=0$ and $u_r=0$.

\begin{proposition}[Structural multipliers]\label{prop:structure}
Let $T(\X)$ denote the stride duration. The apex return map satisfies
\begin{equation}
\dot\theta_{a,k+1}=\dot\theta_{a,k},\qquad \theta_{a,k+1}=\theta_{a,k}+T(\X_k)\,\dot\theta_{a,k}.
\label{eq:pitch_map}
\end{equation}
Every fixed point has $\dot\theta_a^*=0$. At every fixed point, the Jacobian $\mathbf{A}$ has the eigenvalue one with algebraic multiplicity at least two, and its $(\theta_a,\dot\theta_a)$ block is the Jordan block $\left[\begin{smallmatrix}1&T^*\\0&1\end{smallmatrix}\right]$.
\end{proposition}

\begin{proof}
By \eqref{eq:flight} and Proposition~\ref{prop:decoupling}(i), $\ddot\theta=0$ during the whole stride, which gives \eqref{eq:pitch_map}. At $\dot\theta_a=0$, the rows of $\mathbf{A}$ that correspond to $\theta_a$ and $\dot\theta_a$ are $[0,0,1,T^*]$ and $[0,0,0,1]$. Hence $\mathbf{A}$ is block upper triangular, and its spectrum is the spectrum of the $(z_a,\dot x_a)$ block together with a double eigenvalue at one.
\end{proof}

\begin{figure}[!tb]
\centering
\includegraphics[width=\columnwidth]{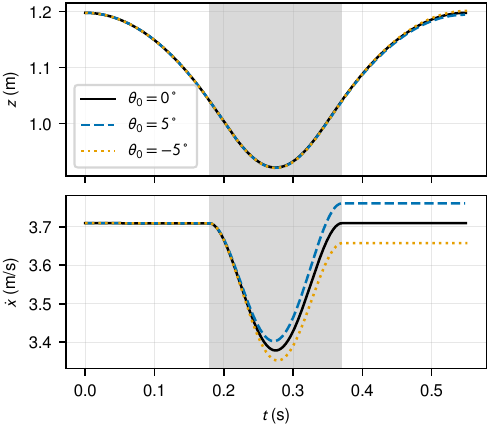}
\caption{One stride of the passive CPP runner from the upright fixed point ($\alpha_\td=109.8^\circ$) with initial pitch $\theta_0\in\{0^\circ,5^\circ,-5^\circ\}$ and zero pitch rate. The pitch stays constant (not shown). Forward lean injects energy and backward lean removes it. The shaded interval marks stance}
\label{fig:lean}
\end{figure}

Proposition~\ref{prop:structure} shows that the passive CPP runner cannot be asymptotically stable. A pitch-rate perturbation persists and the pitch drifts linearly. A pitch perturbation without rate persists as well, and it changes the energy of every following stride (Fig.~\ref{fig:lean}).

\begin{proposition}[Symmetric gaits at upright posture]\label{prop:symmetry}
Let $\theta_a=\dot\theta_a=0$ and $c=0$. Consider a stride whose stance trajectory passes through a state with $x=0$ and $\dot z=0$. Then the stance trajectory is symmetric with respect to this instant, the liftoff state is the mirror image of the touchdown state, and the next apex state equals the current one. The set of such fixed points is defined by a single scalar equation in the $(z_a,\dot x_a)$ plane. It is therefore generically a curve for each $\alpha_\td$, and the $(z_a,\dot x_a)$ block of $\mathbf{A}$ has the eigenvalue one along this curve.
\end{proposition}

\begin{proof}
At $\theta\equiv0$, the leg length $r$ in \eqref{eq:triangle} and the force $N$ in \eqref{eq:N} are even functions of $\thv$. The stance equations \eqref{eq:cpp_radial}--\eqref{eq:cpp_angular} are then reversible with respect to the involution $\mathcal{R}:(x,z,\dot x,\dot z)\mapsto(-x,z,\dot x,-\dot z)$ combined with time reversal \cite{devaney1976reversible,lamb1998time}. A trajectory that intersects the fixed set of $\mathcal{R}$, $\{x=0,\dot z=0\}$, is symmetric with respect to that instant. The leg length is symmetric as well, so the leg returns to $r=r_0$ at the mirrored time. With $c=0$, CPP gives $N\propto F_r$, so liftoff ($N=0$) coincides with $r=r_0$. The liftoff state is therefore $\mathcal{R}$ applied to the touchdown state. The ballistic flight is also reversible, so the next apex has the same height and forward speed. Requiring the trajectory to meet $\{x=0,\dot z=0\}$ imposes one scalar condition on $(z_a,\dot x_a)$. Differentiating $P(\X^*(s))=\X^*(s)$ along the resulting curve gives the unit eigenvalue.
\end{proof}

At upright posture, the $(z_a,\dot x_a)$ block thus has one unit multiplier along the family and one \emph{transverse} multiplier $\lambda_\perp$. The latter determines whether nearby strides converge to the family. This structure parallels the energy-parameterized gait families of SLIP \cite{seyfarth2002movement,GEYER2005315} and the use of symmetry in SLIP return maps \cite{Altendorfer2004}. Here, however, the CPP runner is not conservative, and the families follow from reversibility alone.

For $\theta\neq0$, the symmetry axis $\Phi=0$ of the force field is tilted with respect to the vertical. Strides are then asymmetric and the hip does net work. Figure~\ref{fig:lean} shows one stride from the upright fixed point with $\alpha_\td=109.8^\circ$, $z^*=1.198$\,m and $\dot x^*=3.710$\,m/s, for $\theta=0^\circ$ and $\pm5^\circ$. A forward lean of $5^\circ$ increases the stride energy by 12.7\,J and the apex speed by 0.05\,m/s. A backward lean of $5^\circ$ decreases the energy by 12.6\,J.

\subsection{Periodic gaits of the passive CPP runner}\label{sec:cpp_fp}

Proposition~\ref{prop:symmetry} reduces the search for upright gaits to a scalar equation. For a given pair $(\alpha_\td,z_a)$, we integrate stance until the first instant with $\dot z=0$ and require $x=0$ there. We bracket the roots in $\dot x_a\in[0.05,9]$\,m/s and refine them with Brent's method. We used 117 touchdown angles in $[91^\circ,120^\circ]$ and 96 apex heights in $[1.05,2.0]$\,m. This gives 10{,}977 gaits. Each one is a fixed point of the full return map with a residual below $5\times10^{-14}$, and no pair $(\alpha_\td,z_a)$ yields more than one gait.

\begin{figure*}[!t]
\centering
\includegraphics[width=\textwidth]{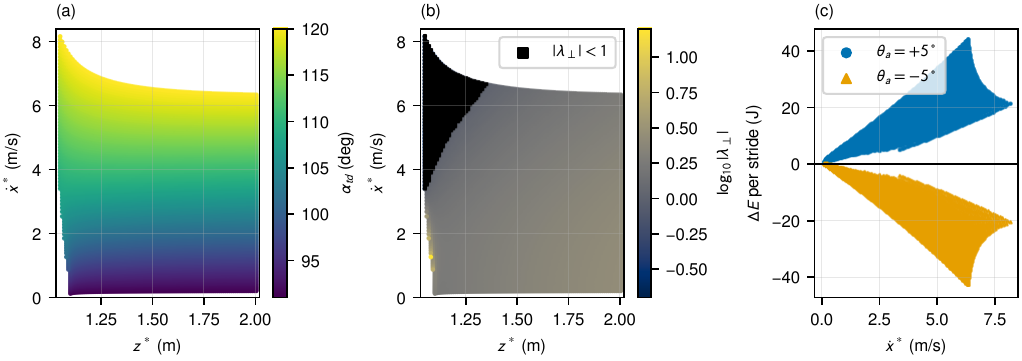}
\caption{Periodic gaits of the passive CPP runner. (a) Gaits in the $(z^*,\dot x^*)$ plane, colored by the touchdown angle; each touchdown angle yields a curve. (b) Transverse multiplier $\lambda_\perp$ on a logarithmic color scale. Black squares have $|\lambda_\perp|<1$. The three structural multipliers equal one in all cases. (c) Energy change during one stride when the gait is started with a constant lean of $\pm5^\circ$}
\label{fig:cpp_fp}
\end{figure*}

Figure~\ref{fig:cpp_fp}(a) shows the gaits. For each $\alpha_\td$ they form a curve, as predicted by Proposition~\ref{prop:symmetry}. Changing $\alpha_\td$ shifts the curve in speed, so the touchdown angle covers a large region of the $(z_a,\dot x_a)$ plane. As predicted by Propositions~\ref{prop:structure} and \ref{prop:symmetry}, the $(z_a,\dot x_a)$ block of the finite-difference Jacobian has one eigenvalue within $5\times10^{-5}$ of one, and the pitch block is the Jordan block of Proposition~\ref{prop:structure}. The eigenvalues of the full $4\times4$ Jacobian deviate from one by up to $2\times10^{-3}$, because the triple unit eigenvalue is not semisimple and finite-difference errors split it. Figure~\ref{fig:cpp_fp}(b) shows the transverse multiplier. It is real and ranges from $-0.20$ to $15.3$. Only 6.0\% of the gaits have $|\lambda_\perp|<1$. They form a region of low apex heights and high speeds ($z^*\in[1.05,1.35]$\,m, $\dot x^*\in[3.4,8.1]$\,m/s, $\alpha_\td\in[108^\circ,120^\circ]$). Even there, the structural multipliers prevent asymptotic stability. Feedback is therefore required. Along the gaits, the peak CPP torque has a median of 45\,N\,m and a maximum of 160\,N\,m, and the peak ratio $|F_x|/F_z$ does not exceed 0.52.

Figure~\ref{fig:cpp_fp}(c) shows that a lean of $+5^\circ$ increases the stride energy at every gait, by 0.3 to 43\,J, and a lean of $-5^\circ$ decreases it. A first-order argument shows when gaits with constant lean can branch off the families. Let $\mathbf{v}_1$ and $\mathbf{v}_\perp$ be eigenvectors of the $(z_a,\dot x_a)$ block for the eigenvalues one and $\lambda_\perp$, and write $\partial P_{z,\dot x}/\partial\theta_a=\xi_1\mathbf{v}_1+\xi_\perp\mathbf{v}_\perp$. A fixed point with a small constant lean $\theta^*$ near a family point $\X_0$ requires $(\lambda_\perp-1)\eta\,\mathbf{v}_\perp+\theta^*(\xi_1\mathbf{v}_1+\xi_\perp\mathbf{v}_\perp)=0$ for the transverse offset $\eta$, and hence $\xi_1(\X_0)=0$. With $\mathbf{v}_1$ oriented toward increasing apex energy, we found $\xi_1>0$ at all 10{,}819 gaits with $|\lambda_\perp-1|>0.05$. A forward lean thus always drives the state along the family toward higher energy, and no branch of gaits with small constant lean emanates from these parts of the families. For the 158 gaits with $|\lambda_\perp-1|\le0.05$, the eigenvectors become nearly parallel and the argument does not apply. A direct search for fixed points on a $70\times70$ grid over $z_a\in[1.02,2.0]$\,m and $\dot x_a\in[0.1,8.5]$\,m/s, for $\theta^*\in\{\pm2^\circ,\pm5^\circ,\pm10^\circ\}$ and $\alpha_\td\in\{95^\circ,100^\circ,\dots,120^\circ\}$, found no gait with $\theta^*\neq0$. Newton runs that diverged to unphysical speeds were discarded.

\begin{remark}\label{rem:aim}
In \cite{karagoz2020aim}, stability was assessed through the largest multiplier modulus of the full $4\times4$ Jacobian. Proposition~\ref{prop:structure} shows that this quantity cannot be smaller than one. The region with $|\lambda_\perp|<1$ in Fig.~\ref{fig:cpp_fp}(b) corresponds to the ``stable'' region reported there. It describes convergence to the gait family, not asymptotic stability.
\end{remark}

%% ===================================================================
\section{Hierarchical gait control}\label{sec:control}
%% ===================================================================

Proposition~\ref{prop:decoupling} suggests a two-level controller. A continuous stance-phase law regulates pitch, and a stride-level law regulates height and speed.

\subsection{Stance-phase pitch control}

We add a PD term to the CPP torque,
\begin{equation}
\tau=\tau_c+\tau_{pd},\qquad \tau_{pd}=-K_p(\theta-\theta^*)-K_d\dot\theta,
\label{eq:pd}
\end{equation}
with the desired pitch $\theta^*$. At the desired posture, $\tau_{pd}=0$ and the GRF passes through the COM. Away from it, $\tau_{pd}$ rotates the GRF and thereby the trunk. The trunk can only be torqued in stance. We use $K_p=999.4$\,N\,m/rad and $K_d=382.7$\,N\,m\,s/rad, which were chosen in \cite{karagoz2020aim} by numerically minimizing the moduli of the pitch-related multipliers at the nominal gait, with both gains bounded by $10^3$. With these gains, the pitch block is nearly deadbeat at the nominal gait (multipliers of modulus $3\times10^{-4}$).

\subsection{Stride-level LQR and closed-loop multipliers}

At each apex, the controller selects the touchdown angle $\alpha_\td$ and the leg damping $c$ for the next stance. Touchdown angle is the classical SLIP input \cite{ghigliazza2005simply,arslan2012reactive}. Radial damping changes the stride energy \cite{secer2018control,ankarali2010stride}. Negative values of $c$ inject energy and must be realized by an actuated leg. Let $\mathbf{u}=[\alpha_\td,\;c]^T$ with nominal value $\mathbf{u}^*=[\alpha_\td^*,\;0]^T$. Linearizing the return map at the fixed point gives
\begin{equation}
\Delta\X_{k+1}=\mathbf{A}\,\Delta\X_k+\mathbf{B}\,\Delta \mathbf{u}_k,
\end{equation}
where $\Delta\X_k=\X_k-\X^*$ and $\Delta \mathbf{u}_k=\mathbf{u}_k-\mathbf{u}^*$. The PD law regulates pitch. The stride-level law therefore acts on $\Delta\X_s=[\Delta z_a,\;\Delta\dot x_a]^T$ through the sub-blocks $\mathbf{A}_s$ and $\mathbf{B}_s$. These sub-blocks do not depend on the PD law, since $\tau_{pd}=0$ when the pitch is not perturbed. At upright gaits, the rows of $\mathbf{A}$ and $\mathbf{B}$ that belong to the pitch states vanish in the first two columns and in $\mathbf{B}$, so the closed-loop spectrum is the union of the spectra of $\mathbf{A}_s-\mathbf{B}_s\mathbf{K}$ and of the pitch block. We use the infinite-horizon LQR
\begin{equation}
\begin{gathered}
\Delta \mathbf{u}_k=-\mathbf{K}\Delta\X_{s,k},\\ \mathbf{K}=(\mathbf{R}+\mathbf{B}_s^T\boldsymbol{\Pi}\mathbf{B}_s)^{-1}\mathbf{B}_s^T\boldsymbol{\Pi}\mathbf{A}_s,
\end{gathered}
\label{eq:lqr}
\end{equation}
where $\boldsymbol{\Pi}$ solves the discrete algebraic Riccati equation with weights $\mathbf{Q}=10\,\mathbf{I}_2$ and $\mathbf{R}=10^{-4}\mathbf{I}_2$, as in \cite{karagoz2020aim}. The weights mix units: $\alpha_\td$ is in radians and $c$ in N\,s/m.

\begin{figure}[!tb]
\centering
\includegraphics[width=\columnwidth]{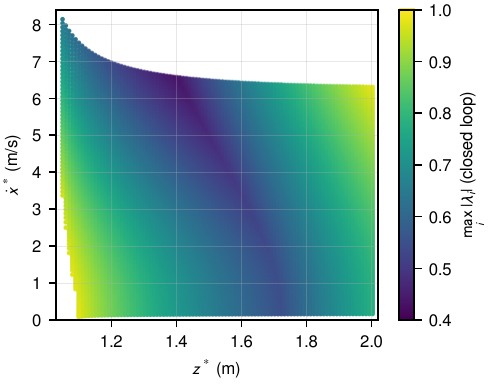}
\caption{Largest closed-loop multiplier modulus with the stance PD law and the stride-level LQR, for all gaits of Fig.~\ref{fig:cpp_fp}}
\label{fig:cl_mult}
\end{figure}

For the nominal gait used below ($\alpha_\td^*=109.8^\circ$, $z^*=1.198$\,m, $\dot x^*=3.710$\,m/s), the open-loop Jacobian has $\lambda_\perp=1.36$, two unit multipliers in the Jordan block of Proposition~\ref{prop:structure} with $T^*=0.55$\,s, and the family multiplier at one. The LQR gain is
\begin{equation}
\mathbf{K}=\begin{bmatrix}0.0997\,\mathrm{rad/m}&-0.2012\,\mathrm{rad\,s/m}\\-250.3\,\mathrm{N\,s/m^2}&-104.6\,\mathrm{N\,s^2/m^2}\end{bmatrix}.
\end{equation}
The $(z_a,\dot x_a)$ block then has the closed-loop multipliers 0.74 and 0. The slower one is due to the weight on $c$: reducing it to $10^{-8}$ makes the block nearly deadbeat (multipliers below $10^{-3}$). We computed $\mathbf{K}$ and the closed-loop multipliers for every gait of Fig.~\ref{fig:cpp_fp}. All closed-loop multipliers lie inside the unit circle (Fig.~\ref{fig:cl_mult}). The largest modulus ranges from 0.43 to 0.976, with a median of 0.67. In 58\% of the gaits the slowest mode belongs to the $(z_a,\dot x_a)$ block, whose slowest multiplier stems from the input weighting. In the others it belongs to the pitch block. The pitch gains were tuned at the nominal gait, and the pitch block becomes slow for gaits with short stance and long flight, such as $z^*=2.0$\,m and $\dot x^*=6.3$\,m/s (stance 0.18\,s, stride 1.09\,s).

\subsection{Torque limits, recovery and basins}\label{sec:limits}

The PD term can demand large torques after pitch disturbances. By \eqref{eq:grf}, a large torque also tilts the GRF and can violate the friction cone. For a given state and leg force, the conditions $|F_x|\le\mu F_z$ and $F_z\ge0$ are linear in $\tau$. They define an interval $[\tau_{\min},\tau_{\max}]$, which we compute at every integration step. We clip the total torque $\tau_c+\tau_{pd}$ first to this interval and then to $|\tau|\le\bar\tau$. If the interval is empty, which happens only briefly near touchdown and liftoff, we use its midpoint. When the two limits conflict, the torque limit wins, so the friction ratio can slightly exceed $\mu$; in the basin runs it stayed below 0.89. We use $\mu=0.8$ and $\bar\tau=300$\,N\,m (3.75\,N\,m/kg) as illustrative limits. The CPP torque itself stays well inside these limits along all gaits (Section~\ref{sec:cpp_fp}). Clipping therefore acts only after disturbances.

\begin{figure*}[!t]
\centering
\includegraphics[width=\textwidth]{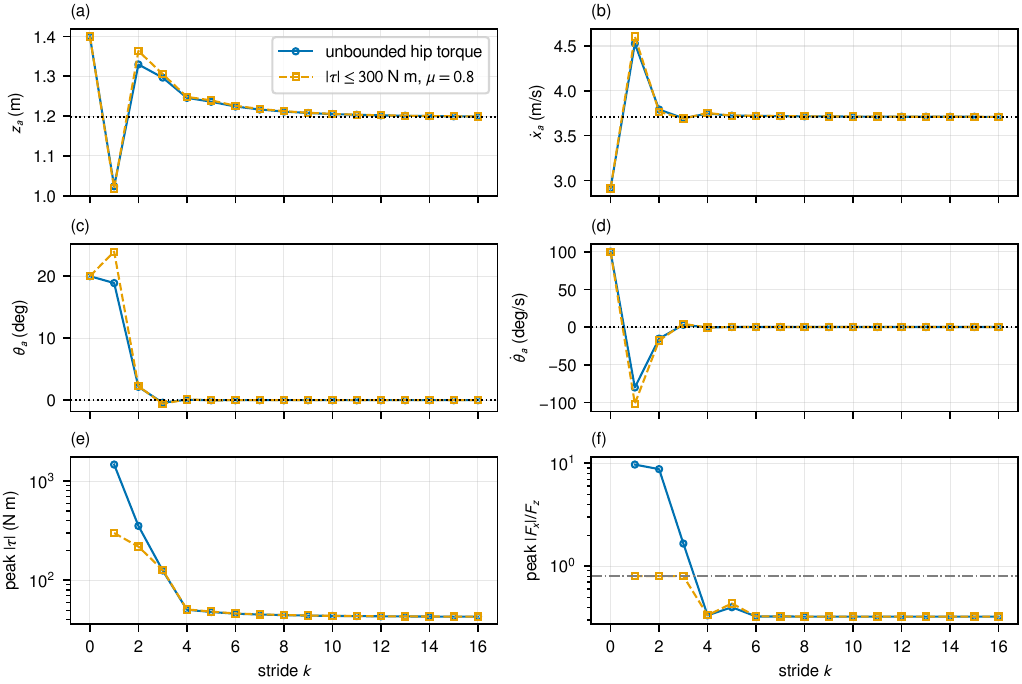}
\caption{Recovery from $\X_0=[1.4\,\mathrm{m},\,2.91\,\mathrm{m/s},\,20^\circ,\,100^\circ/\mathrm{s}]^T$ under the PD--LQR controller, with unbounded hip torque and with $|\tau|\le300$\,N\,m and $\mu=0.8$. (a)--(d) Apex states; dotted lines mark the nominal gait. (e) Peak hip torque and (f) peak friction ratio in each stance}
\label{fig:cl_example}
\end{figure*}

Figure~\ref{fig:cl_example} shows a recovery from a large disturbance. The runner starts at $z_a=1.4$\,m and $\dot x_a=2.91$\,m/s with $\theta_a=20^\circ$ and $\dot\theta_a=100^\circ$/s. In the first stride, the controller selects $\alpha_\td=99.4^\circ$ and $c=-33$\,N\,s/m, which injects energy. The pitch falls below $1^\circ$ from the third apex on. Height and speed are within 1\% of the nominal gait from the ninth apex on. Without limits, the first stance requires a hip torque of 1{,}470\,N\,m and a friction ratio $|F_x|/F_z$ of 9.7, which is not realistic. With the limits of Section~\ref{sec:limits}, the first two apex states differ (for example, the pitch at the first apex is $24.0^\circ$ instead of $18.9^\circ$), but the convergence is equally fast.

To estimate basins of attraction, we simulated the closed loop from a $100\times100$ grid of initial apex heights and speeds around the nominal gait, for three initial pitch conditions: upright, $\theta_0=20^\circ$, and $\dot\theta_0=100^\circ$/s. A run counts as convergent if no failure occurs within 30 strides and the final apex state is within 1\% of $\X^*$ in relative norm. All convergent runs met this criterion within 6\,s. Failures are a hip below the ground, a pitch rate above $120^\circ$/s, a leg shorter than 0.2\,m, or a stride without touchdown or apex. The stride-level inputs are not bounded; in convergent runs, $\alpha_\td$ ranged from about $88^\circ$ to $139^\circ$ and $|c|$ reached about 330\,N\,s/m. For comparability with \cite{karagoz2020aim}, the grid spans $z_0\in[0.8,2.0]$\,m, although initial heights below about 1.04\,m do not allow touchdown at $\alpha_\td^*$.

\begin{figure*}[!t]
\centering
\includegraphics[width=\textwidth]{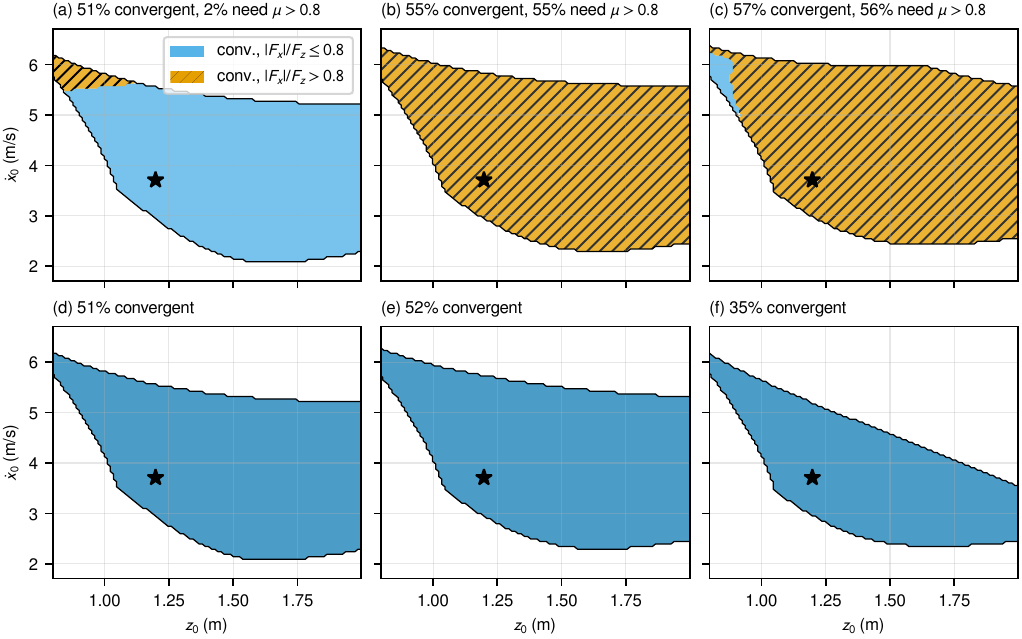}
\caption{Basins of attraction of the PD--LQR controller in the plane of initial apex height and speed. The star marks the nominal gait. Top row: unbounded hip torque; hatched regions converge only with a peak friction ratio above 0.8. Bottom row: hip torque and friction limits of Section~\ref{sec:limits}. Columns: upright start, initial pitch $20^\circ$, initial pitch rate $100^\circ$/s}
\label{fig:basins}
\end{figure*}

Figure~\ref{fig:basins} shows the results. With unbounded torque, the controller recovers from 51\%, 55\% and 57\% of the tested region for the three cases. However, almost all recoveries from pitch disturbances require $|F_x|/F_z>0.8$ at some instant. With the limits, the basins cover 51\%, 52\% and 35\% of the region. Restricted to $z_0\ge1.05$\,m, the fractions are 58\%, 62\% and 65\% without limits and 58\%, 59\% and 38\% with limits. The upright basin is unchanged by the limits. A lean of $20^\circ$ is recovered from almost the same region. A pitch rate of $100^\circ$/s mainly removes recoveries from high speeds. Without limits, the pitch-disturbed basins are slightly larger than the upright one at high initial speeds. There, the unbounded LQR brakes strongly in the first stride (for example, $\alpha_\td=129^\circ$ and $c=254$\,N\,s/m from $z_0=1.47$\,m and $\dot x_0=5.5$\,m/s), and the energy injected by forward lean compensates part of this overcorrection.

%% ===================================================================
\section{Posture-invariant template embedding}\label{sec:embedding}
%% ===================================================================

The CPP stance map has no closed form because the radial force $N$ in \eqref{eq:cpp_radial} is nonlinear and posture dependent. We now use the leg actuator force $u_r$ in \eqref{eq:Fr} to impose a chosen radial force law in virtual-leg coordinates. This is a PFL of the radial dynamics \cite{Piovan2016,Hamzacebi2017}, applied to the virtual leg rather than the physical leg.

\begin{figure}[!tb]
\centering
\includegraphics[width=0.8\columnwidth]{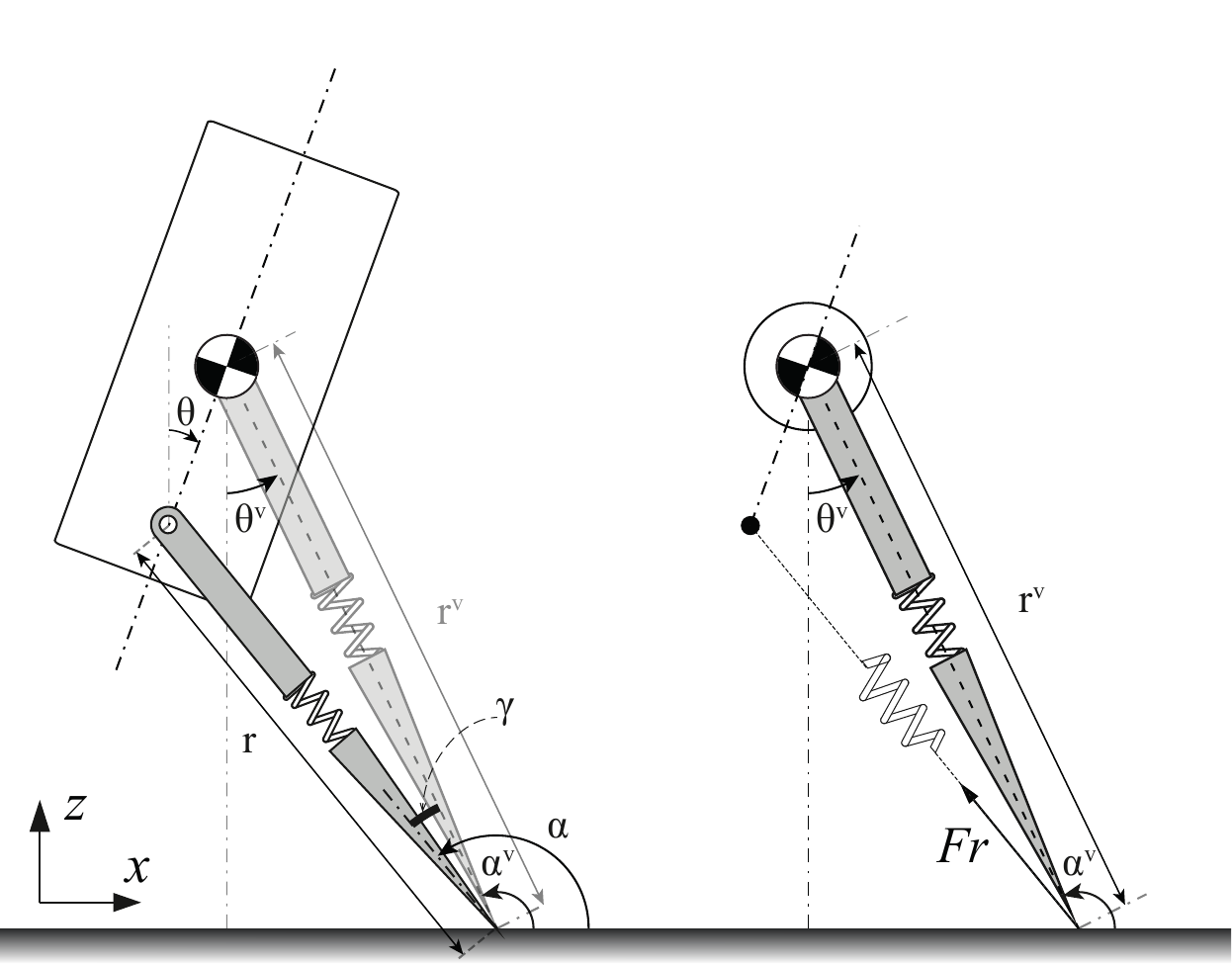}
\caption{Physical leg $(r,\alpha)$, virtual leg $(\rv,\thv)$ and the angle $\gamma$ between them. Under CPP and the embedding of Section~\ref{sec:embedding}, the COM moves as a point mass on a virtual leg with a prescribed radial force law}
\label{fig:virtualleg}
\end{figure}

Let $N_{\mathrm{des}}(\rv,\thv,\dot r^{v},\dot\theta^{v})$ be a desired GRF magnitude. Using \eqref{eq:N} and \eqref{eq:Fr}, the leg actuator force
\begin{equation}
u_r=\frac{\rv-\dhip\cos\Phi}{r}\,N_{\mathrm{des}}+k(r-r_0)+c\,\dot r
\label{eq:ur}
\end{equation}
yields $N=N_{\mathrm{des}}$. The first term sets the GRF, and the last two terms cancel the physical spring and damper. The hip torque remains the CPP torque \eqref{eq:tau_cpp}, now evaluated with the total leg force $F_r$. The rest length of the virtual leg is the virtual leg length at touchdown. From the touchdown geometry,
\begin{equation}
\begin{gathered}
\rvz=\sqrt{r_0^2+\dhip^2+2r_0\dhip\sin(\alpha_\td+\theta_\td)},\\
\tan\thv_\td=-\frac{r_0\cos\alpha_\td+\dhip\sin\theta_\td}{r_0\sin\alpha_\td+\dhip\cos\theta_\td},
\end{gathered}
\label{eq:td_geometry}
\end{equation}
where $\theta_\td$ is the pitch at touchdown.

\begin{proposition}[Posture-invariant embedding]\label{prop:embedding}
Let $N_{\mathrm{des}}$ depend only on the virtual-leg states and on constants fixed at touchdown. Under $\tau=\tau_c$ and the actuator force \eqref{eq:ur}, the stance dynamics are
\begin{equation}
\begin{gathered}
m\big(\ddot r^{v}-\rv(\dot\theta^{v})^2+g\cos\thv\big)=N_{\mathrm{des}},\\ \ddot\theta^{v}=\frac{g\sin\thv-2\dot r^{v}\dot\theta^{v}}{\rv},\qquad \ddot\theta=0,
\end{gathered}
\label{eq:embedded}
\end{equation}
and liftoff occurs when $N_{\mathrm{des}}=0$. Consequently, the COM stance trajectory depends on the pitch angle and pitch rate only through the touchdown pair $(\rvz,\thv_\td)$ in \eqref{eq:td_geometry}.
\end{proposition}

\begin{proof}
Substituting \eqref{eq:ur} into \eqref{eq:Fr} and \eqref{eq:N} gives $N=N_{\mathrm{des}}$. Equations \eqref{eq:cpp_radial} and \eqref{eq:cpp_angular} then give \eqref{eq:embedded}, and $\ddot\theta=0$ follows from Proposition~\ref{prop:decoupling}. None of these equations contains $\theta$ or $\dot\theta$. The vertical GRF is $N\cos\thv$, so liftoff occurs at $N=0$. The initial conditions depend on $\theta$ only through the touchdown geometry.
\end{proof}

Proposition~\ref{prop:embedding} has two consequences. First, any constant pitch can be held during running without affecting the COM stance dynamics. Pitch enters the stride map only through the touchdown geometry, which acts as a second geometric input besides $\alpha_\td$. Second, the choice of $N_{\mathrm{des}}$ is free. We verified \eqref{eq:ur}--\eqref{eq:embedded} against the full model to machine precision.

We consider two templates. The \emph{SLIP template} $N_{\mathrm{des}}=k^v(\rvz-\rv)$ turns the COM motion into that of a SLIP with stiffness $k^v$ and rest length $\rvz$. The TSLIP with CPP then inherits all SLIP results, with a rest length that depends on posture through \eqref{eq:td_geometry}. Its stance map, however, is not integrable.

The \emph{linear template} removes the centripetal and gravitational terms from the radial dynamics:
\begin{equation}
\begin{gathered}
N_{\mathrm{des}}=k^v(\rvz-\rv)+m\big(g\cos\thv-\rv(\dot\theta^{v})^2\big),\\ \ddot r^{v}=\omega_0^2\big(\rvz-\rv\big),\qquad \omega_0=\sqrt{k^v/m}.
\end{gathered}
\label{eq:template}
\end{equation}
The radial motion is then a harmonic oscillation. The angular motion keeps gravity and is not integrable. This template was used in \cite{sever2022thesis}. The rest of the paper focuses on it, and we compare it with the SLIP template in Section~\ref{sec:prediction}.

\label{sec:admissible}
The ground can only push, so the embedding requires $N_{\mathrm{des}}\ge0$ during stance. For the SLIP template, $N_{\mathrm{des}}$ is zero at touchdown and positive while the virtual leg is compressed. For the linear template, $N_{\mathrm{des}}$ jumps at touchdown from zero to
\begin{equation}
\Ntd=m\big(g\cos\thv_\td-\rvz(\dot\theta^{v}_\td)^2\big).
\label{eq:Ntd}
\end{equation}
The leg actuator must produce the corresponding force step $u_r=\Ntd(\rvz-\dhip\cos\Phi_\td)/r_0$ while the physical spring is still unloaded. We call a stride \emph{admissible} if $\Ntd\ge0$, that is, if the tangential COM speed at touchdown does not exceed $\sqrt{g\rvz\cos\thv_\td}$. In the simulations, liftoff is detected when the vertical GRF returns to zero after loading. Section~\ref{sec:natural} shows that the step vanishes at all natural gaits.

%% ===================================================================
\section{Approximate analytic apex return map}\label{sec:approx}
%% ===================================================================

\begin{figure}[!tb]
\centering
\includegraphics[width=\columnwidth]{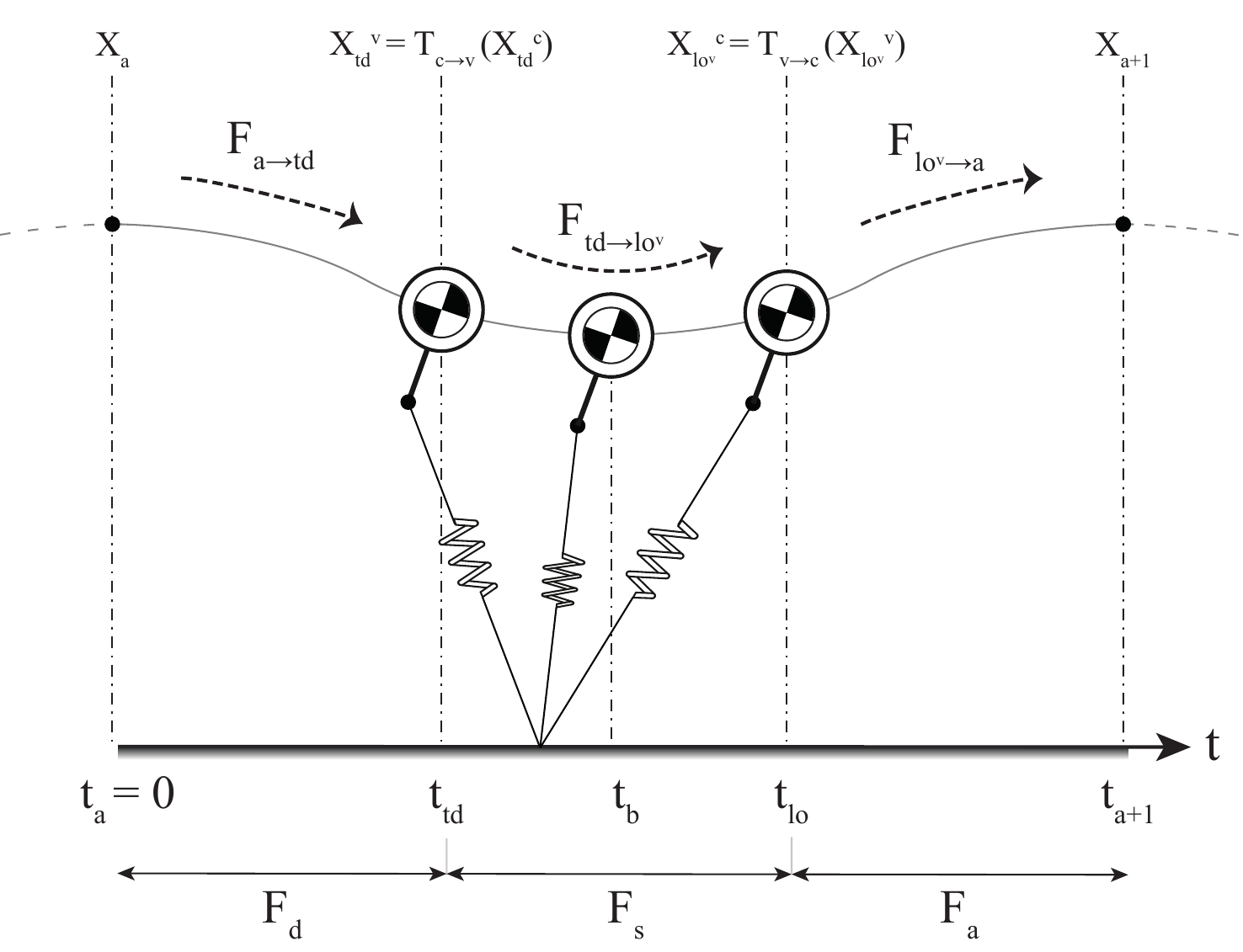}
\caption{Decomposition of the apex return map into descent, stance and ascent maps with coordinate changes between Cartesian and virtual-leg coordinates}
\label{fig:apexmap}
\end{figure}

We decompose the apex return map of the runner with the linear template as (Fig.~\ref{fig:apexmap})
\begin{equation}
P=F_a\circ T_{v\to c}\circ F_s\circ T_{c\to v}\circ F_d,
\end{equation}
where $F_d$, $F_s$ and $F_a$ are the descent, stance and ascent maps, and $T_{c\to v}$ and $T_{v\to c}$ are coordinate changes (Appendix~\ref{app:transform}). Flight is exact. Only $F_s$ is approximated.

\subsection{Descent and ascent}

From the apex, the COM falls ballistically and the pitch evolves as $\theta(t)=\theta_a+\dot\theta_a t$. The touchdown time $t_\td$ is the smallest positive root of
\begin{equation}
z_a-\tfrac12 g t^2-\dhip\cos(\theta_a+\dot\theta_a t)-r_0\sin\alpha_\td=0.
\label{eq:ttd}
\end{equation}
For $\dot\theta_a=0$, $t_\td=\sqrt{2(z_a-\dhip\cos\theta_a-r_0\sin\alpha_\td)/g}$. Otherwise, \eqref{eq:ttd} is a scalar root-finding problem. The ascent from liftoff is ballistic, and the next apex occurs $\dot z_\lo/g$ after liftoff. The pitch rate is unchanged throughout the stride.

\subsection{Stance: exact radius and exact sweep}

We measure time from touchdown. With $\rv(0)=\rvz$, the radial solution of \eqref{eq:template} is exact:
\begin{equation}
\begin{gathered}
\rv(t)=\rvz+a\sin\omega_0 t,\qquad \dot r^{v}(t)=\dot r^{v}_\td\cos\omega_0 t,\\ a=\dot r^{v}_\td/\omega_0<0.
\end{gathered}
\label{eq:rv}
\end{equation}
Let $h=(\rv)^2\dot\theta^{v}$ denote the specific angular momentum about the toe. By \eqref{eq:cpp_angular}, $\dot h=g\rv\sin\thv$. For a constant $h$, the sweep angle follows from $\dot\theta^{v}=h/(\rv)^2$. In SLIP approximations, $\rv(t)$ itself is only approximate, and $1/(\rv)^2$ is expanded to first order \cite{GEYER2005315,Saranli2010}. Here $\rv(t)$ is exact, and the sweep can be integrated exactly:
\begin{equation}
\thv(t)=\thv_\td+\frac{h}{\omega_0}\,\mathcal{J}(\omega_0t;\rvz,a),\qquad \dot\theta^{v}(t)=\frac{h}{\rv(t)^2},
\label{eq:sweep}
\end{equation}
with
\begin{align}
\mathcal{J}(\varphi;\rho,a)&=\int_0^{\varphi}\frac{d\phi}{(\rho+a\sin\phi)^2}\nonumber\\
&=\frac{1}{\rho^2-a^2}\Big(\frac{a\cos\varphi}{\rho+a\sin\varphi}-\frac{a}{\rho}+\rho\,\mathcal{I}(\varphi)\Big),\label{eq:J}\\
\mathcal{I}(\varphi)&=\frac{2}{\ell}\arctan\frac{\rho\tan(\varphi/2)+a}{\ell}\nonumber\\
&\quad-\frac{2}{\ell}\arctan\frac{a}{\ell},\label{eq:K}\\
\ell&=\sqrt{\rho^2-a^2}.\nonumber
\end{align}
The function $\mathcal{I}$ is the integral of $1/(\rho+a\sin\phi)$ and must be continuous. At $\varphi=\pi$, the first arctangent is replaced by its limit $\pi/2$, and for $\varphi\in(\pi,3\pi)$, $\pi$ is added to it. The formula requires $|a|<\rvz$, which holds for any compression smaller than the rest length. Without gravity correction, $h=(\rvz)^2\dot\theta^{v}_\td$.

By Proposition~\ref{prop:embedding}, liftoff occurs when $N_{\mathrm{des}}=0$, that is, when $\ddot r^{v}=\rv(\dot\theta^{v})^2-g\cos\thv$. We evaluate the right-hand side at the symmetric estimate $t=\pi/\omega_0$, where $\rv=\rvz$:
\begin{equation}
\ddot r^{v}_\lo\approx\rvz(\dot\theta^{v}_\td)^2-g\cos\!\Big(\thv_\td+\frac{h}{\omega_0}\mathcal{J}(\pi;\rvz,a)\Big).
\end{equation}
Equating this value to $\ddot r^{v}(t)=-\omega_0\dot r^{v}_\td\sin\omega_0 t$ from \eqref{eq:rv} gives
\begin{equation}
t_s=\frac{1}{\omega_0}\Big(\pi+\arcsin\frac{\ddot r^{v}_\lo}{\omega_0\dot r^{v}_\td}\Big).
\label{eq:tlo}
\end{equation}
Since $\dot r^{v}_\td<0$ and typically $\ddot r^{v}_\lo<0$, the stance lasts slightly longer than half a period.

\subsection{Gravity correction}

Gravity changes the angular momentum during stance. For asymmetric strides, the change is large and mainly affects the liftoff velocity. Let $\Gamma=g\rv\sin\thv$ denote the specific gravitational moment about the toe, with the values $\Gamma_\td=g\rvz\sin\thv_\td$ and $\Gamma_\lo=g\rv_\lo\sin\thv_\lo$ at touchdown and liftoff. We approximate $\Gamma$ as linear in time. Then
\begin{equation}
h(t)=h_\td+\Gamma_\td t+(\Gamma_\lo-\Gamma_\td)\frac{t^2}{2t_s}.
\end{equation}
The \emph{linear-moment correction} (GC-L) uses the stance average of $h(t)$ for the sweep and its end value for the liftoff rate:
\begin{equation}
\begin{gathered}
\bar h=h_\td+t_s\Big(\frac{\Gamma_\td}{3}+\frac{\Gamma_\lo}{6}\Big),\\
h_\lo=h_\td+\frac{t_s}{2}(\Gamma_\td+\Gamma_\lo),
\end{gathered}
\label{eq:gcl}
\end{equation}
so that $\thv_\lo=\thv_\td+\bar h\,\mathcal{J}(\omega_0t_s;\rvz,a)/\omega_0$ and $\dot\theta^{v}_\lo=h_\lo/(\rv_\lo)^2$. Using the average $\bar h$ in the sweep neglects the correlation between $h(t)$ and $1/\rv(t)^2$. Since $\Gamma_\lo$ depends on $\thv_\lo$, we start from the uncorrected sweep and iterate \eqref{eq:gcl} three times. The liftoff radius and radial rate follow exactly from \eqref{eq:rv} at $t=t_s$. The complete map uses only elementary functions, apart from the scalar root in \eqref{eq:ttd} when $\dot\theta_a\neq0$.

Earlier corrections replace $h$ by a single constant for both purposes. The mean-momentum correction (GC-M) of \cite{Saranli2010} averages the touchdown and liftoff momenta, and the total-momentum correction (GC-T) of \cite{Arslan2009} adds the whole gravitational impulse. GC-L is a small change of these corrections. It averages the moment instead of the momentum, and it uses different momenta for the sweep and for the rate. Combined with the first-order expansion of $1/(\rv)^2$, GC-M and GC-T were used in \cite{sever2022thesis}. We use them, and the uncorrected expansion (NC), as baselines. Appendix~\ref{app:angular} lists their formulas. For the SLIP template, we use the lossless approximation of \cite{Saranli2010} with its energy correction, which fixes the liftoff speed by energy conservation.

%% ===================================================================
\section{Prediction performance}\label{sec:prediction}
%% ===================================================================

\subsection{Protocol}

We compare the approximate maps with the numerical return map of the full TSLIP model \eqref{eq:stance} under CPP and the embedding force \eqref{eq:ur}. Following \cite{Arslan2009,Saranli2010}, we use the percentage prediction error
\begin{equation}
\PE(\mathbf{y})=100\,\frac{\|\mathbf{y}_{\mathrm{num}}-\mathbf{y}_{\mathrm{approx}}\|_2}{\|\mathbf{y}_{\mathrm{num}}\|_2}
\label{eq:pe}
\end{equation}
for the liftoff position $\mathbf{p}_\lo=[x_\lo,z_\lo]$, the liftoff velocity $\dot{\mathbf{p}}_\lo=[\dot x_\lo,\dot z_\lo]$, the apex position $\mathbf{p}_a=[x_a,z_a]$, the apex height $z_a$, the apex energy $E_a$ and the stance duration $t_s$. Positions are measured from the toe. We use the liftoff velocity instead of the apex velocity, which can vanish.

Table~\ref{tab:ranges} lists the tested grid, which follows \cite{sever2022thesis}. The stiffness range covers human running \cite{Arampatzis}. A stride is excluded if the next apex is lower than $r_0$, if the apex lies less than 0.1\,m ahead of the toe, if $\dot z_\lo\le0$, or if the virtual leg becomes shorter than $0.75\,r_0$, as in \cite{sever2022thesis} (compare the compression limits in \cite{Arslan2009,Saranli2010}). For the linear template, we also exclude inadmissible strides with $\Ntd<0$ (Section~\ref{sec:admissible}).

\begin{table}[!tb]
\caption{Apex states and parameters used to evaluate the approximate maps. Grid sizes are given in parentheses. Set B uses five values each for $z_a$, $\dot x_a$ and $k^v$}\label{tab:ranges}
\begin{tabular}{@{}lll@{}}
\toprule
Quantity & Range & Grid\\
\midrule
$z_a$ & $[1.1,\,1.5]$\,m & (10)\\
$\dot x_a$ & $[0,\,8]$\,m/s & (10)\\
$\theta_a$ & $\{-10^\circ,0^\circ,10^\circ\}$ & (3)\\
$\dot\theta_a$ & set A: $0$; set B: $\{7.5,\,15\}^\circ$/s & (1); (2)\\
$\alpha_\td$ & $[90.5^\circ,124^\circ]$ & (19)\\
$k^v$ ($\omega_0$) & $[8,36]$\,kN/m ($[10.0,21.2]$\,rad/s) & (10)\\
\botrule
\end{tabular}
\end{table}

\subsection{Results}

For the linear template, the validity filter retained 14{,}616 of the 57{,}000 strides in set A (25.6\%) and 3{,}093 of the 14{,}250 strides in set B (21.7\%). Of these, 10{,}846 (74.2\%) and 2{,}240 (72.4\%) are admissible. Inadmissible strides occur at high speeds. All valid strides with $\dot x_a\ge6.2$\,m/s and 96\% of those with $\dot x_a=5.3$\,m/s are inadmissible, whereas all strides with $\dot x_a\le2.7$\,m/s are admissible. For the SLIP template, 19{,}671 strides of set A (34.5\%) and 4{,}217 of set B (29.6\%) are valid.

\begin{figure*}[!t]
\centering
\includegraphics[width=\textwidth]{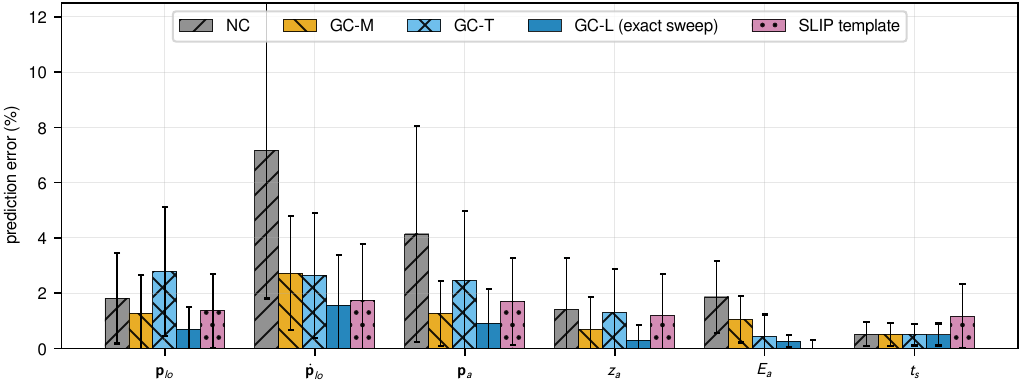}
\caption{Mean prediction errors for set A; whiskers show one standard deviation. The bars of each quantity are ordered as in the legend}
\label{fig:pe_summary}
\end{figure*}

Table~\ref{tab:pe} and Fig.~\ref{fig:pe_summary} summarize the prediction errors. In set A, the exact sweep with the linear-moment correction (GC-L) is the most accurate map for all quantities except the apex energy; for the stance duration it ties with GC-T. In set A, its mean errors are 0.70\% for the liftoff position, 1.56\% for the liftoff velocity and 0.91\% for the apex position. The best earlier correction, GC-M, gives 1.28\%, 2.73\% and 1.27\%, and the uncorrected expansion 1.82\%, 7.16\% and 4.14\%. The SLIP-template approximation gives 1.36\%, 1.75\% and 1.69\%. Its energy correction makes the apex energy almost exact, but its stance duration error is twice as large. On the 10{,}275 strides that are valid for both templates, GC-L gives 1.34\% for the liftoff velocity and 0.81\% for the apex position, and the SLIP-template approximation 1.86\% and 1.67\%. The gravity-corrected variant of the SLIP approximation was less accurate for the velocity (3.03\%). The largest GC-L errors are 7.4\%, 19.3\% and 14.6\%. They occur for strongly asymmetric strides ($|\sigma|>20^\circ$, see below), mostly with the softest virtual leg. Including the inadmissible strides changes the GC-L means to 0.65\%, 1.27\% and 0.76\%.

\begin{table*}[!t]
\caption{Percentage prediction errors (mean $\pm$ standard deviation) of the approximate maps on admissible strides. NC, GC-M, GC-T: first-order expansion without correction, with mean-momentum and with total-momentum correction. GC-L: exact sweep with the linear-moment correction \eqref{eq:gcl}. SLIP: SLIP template with the approximation of \cite{Saranli2010} and energy correction. Strides for which a map predicts no flight phase are excluded for that map (set A: 0, 19, 68, 25 and 56 strides for the five maps; set B: 0, 8, 10, 6 and 4)}\label{tab:pe}
\footnotesize
\begin{tabular}{@{}lccccc@{}}
\toprule
& NC & GC-M & GC-T & GC-L & SLIP\\
\midrule
\multicolumn{6}{@{}l}{Set A ($\dot\theta_a=0$)}\\
$\mathbf{p}_\lo$ & 1.82$\pm$1.64 & 1.28$\pm$1.36 & 2.78$\pm$2.35 & \textbf{0.70}$\pm$0.79 & 1.36$\pm$1.34\\
$\dot{\mathbf{p}}_\lo$ & 7.16$\pm$5.35 & 2.73$\pm$2.05 & 2.63$\pm$2.26 & \textbf{1.56}$\pm$1.83 & 1.75$\pm$2.04\\
$\mathbf{p}_a$ & 4.14$\pm$3.90 & 1.27$\pm$1.18 & 2.47$\pm$2.49 & \textbf{0.91}$\pm$1.22 & 1.69$\pm$1.57\\
$z_a$ & 1.40$\pm$1.89 & 0.70$\pm$1.17 & 1.31$\pm$1.57 & \textbf{0.30}$\pm$0.55 & 1.20$\pm$1.48\\
$E_a$ & 1.86$\pm$1.31 & 1.05$\pm$0.83 & 0.45$\pm$0.78 & 0.27$\pm$0.22 & \textbf{0.01}$\pm$0.28\\
$t_s$ & 0.51$\pm$0.44 & 0.51$\pm$0.42 & 0.50$\pm$0.40 & \textbf{0.50}$\pm$0.40 & 1.16$\pm$1.16\\
\midrule
\multicolumn{6}{@{}l}{Set B ($\dot\theta_a\neq0$)}\\
$\mathbf{p}_\lo$ & 1.96$\pm$1.76 & 1.51$\pm$1.46 & 2.99$\pm$2.33 & \textbf{0.75}$\pm$0.88 & 1.31$\pm$1.27\\
$\dot{\mathbf{p}}_\lo$ & 7.37$\pm$5.18 & 2.89$\pm$2.16 & 2.88$\pm$2.21 & 1.64$\pm$1.90 & \textbf{1.63}$\pm$1.87\\
$\mathbf{p}_a$ & 4.36$\pm$3.75 & 1.45$\pm$1.26 & 2.62$\pm$2.40 & \textbf{0.95}$\pm$1.22 & 1.66$\pm$1.45\\
$z_a$ & 1.56$\pm$2.17 & 0.86$\pm$1.23 & 1.52$\pm$1.59 & \textbf{0.34}$\pm$0.71 & 1.20$\pm$1.47\\
$E_a$ & 1.95$\pm$1.32 & 1.10$\pm$0.90 & 0.49$\pm$0.91 & 0.28$\pm$0.21 & \textbf{0.00}$\pm$0.15\\
$t_s$ & 0.53$\pm$0.42 & 0.53$\pm$0.41 & 0.53$\pm$0.41 & \textbf{0.52}$\pm$0.40 & 1.20$\pm$1.18\\
\botrule
\end{tabular}
\end{table*}

The results for set B are close to those of set A; there, the SLIP-template approximation is as accurate as GC-L for the liftoff velocity. A nonzero pitch rate therefore does not degrade the prediction, as expected from Proposition~\ref{prop:embedding}. The pitch angle has no measurable effect either. With GC-L, the mean liftoff velocity error is 1.56\% at $\theta_a=-10^\circ$, $0^\circ$ and $10^\circ$, and the mean apex position errors are 0.95\%, 0.91\% and 0.88\%.

\begin{figure*}[!t]
\centering
\includegraphics[width=\textwidth]{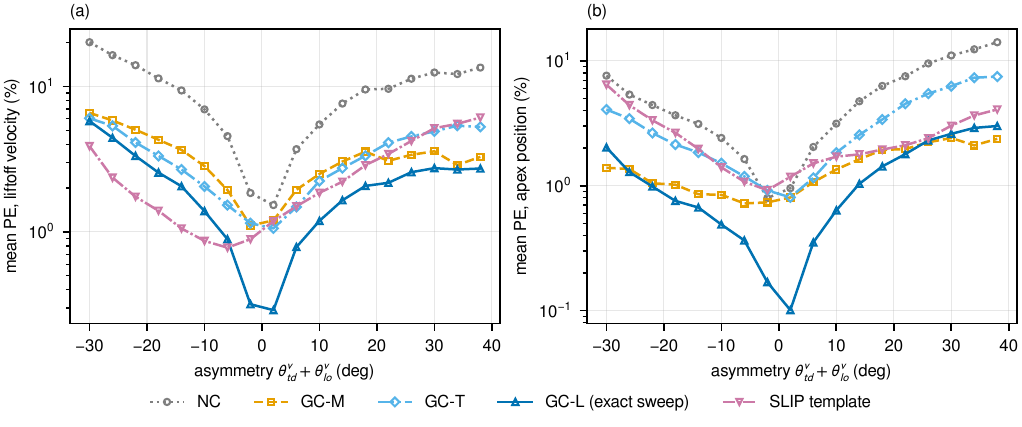}
\caption{Mean prediction errors of (a) the liftoff velocity and (b) the apex position versus the stride asymmetry $\thv_\td+\thv_\lo$ for set A (bins of $4^\circ$ with at least 20 strides)}
\label{fig:pe_asym}
\end{figure*}

\begin{figure*}[!t]
\centering
\includegraphics[width=\textwidth]{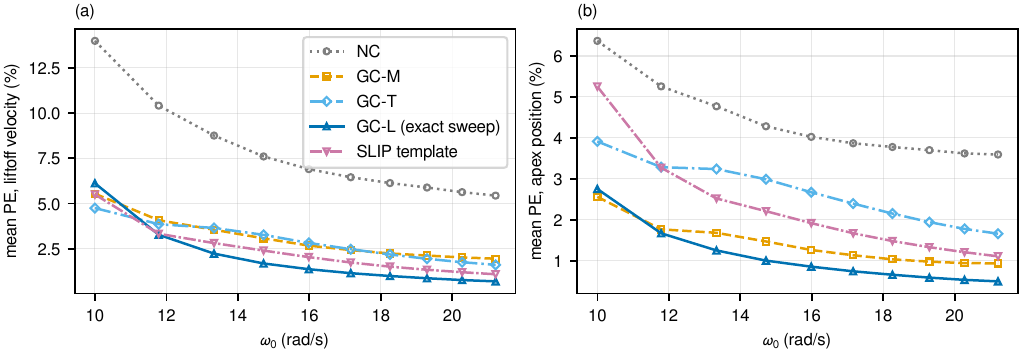}
\caption{Mean prediction errors of (a) the liftoff velocity and (b) the apex position versus the natural frequency $\omega_0$ of the template for set A}
\label{fig:pe_omega}
\end{figure*}

Figure~\ref{fig:pe_asym} shows the errors as a function of the stride asymmetry $\sigma=\thv_\td+\thv_\lo$, which vanishes for symmetric strides. All maps are most accurate near $\sigma=0$. There, GC-L is almost exact: for $|\sigma|<5^\circ$, its mean liftoff velocity and apex position errors are 0.4\% and 0.2\% or less. For the apex position, GC-L is the most accurate map for $|\sigma|<25^\circ$, which contains 87\% of the admissible strides. For the liftoff velocity, the SLIP-template approximation is more accurate for $\sigma<-5^\circ$. Its energy correction fixes the speed magnitude, which the linear template cannot do because it is not conservative. Figure~\ref{fig:pe_omega} shows that the errors decrease with the template frequency $\omega_0$ for all maps. Stiffer virtual legs compress less, and strides become shorter and less affected by gravity. GC-L is the most accurate map for all $\omega_0\ge11.8$\,rad/s.

On the admissible strides of set A, the GRF step $\Ntd$ at touchdown has a median of 443\,N ($0.57\,mg$), a 95th percentile of 739\,N and a maximum of 759\,N. The peak actuator force $|u_r|$ is similar for both templates (medians of 1.30 and 1.33\,kN). The peak hip torque has medians of 74\,N\,m (linear template) and 101\,N\,m (SLIP template), and the peak friction ratio medians of 0.35 and 0.43.

%% ===================================================================
\section{Natural gaits of the embedded runner}\label{sec:natural}
%% ===================================================================

We now consider the passive embedded runner with the linear template: CPP, the actuator force \eqref{eq:ur}, a constant pitch and no PD term. By Proposition~\ref{prop:embedding}, the touchdown geometry $(\rvz,\thv_\td)$ is fixed for given $\alpha_\td$ and $\theta^*$. The touchdown state is then determined by the touchdown rates $(\dot r^{v}_\td,\dot\theta^{v}_\td)$, which are in one-to-one correspondence with the apex state $(z_a,\dot x_a)$. We call its periodic gaits natural gaits. They have a remarkable structure.

\begin{proposition}[Natural gaits]\label{prop:natural}
Consider the passive embedded runner with the linear template \eqref{eq:template}, constant pitch, zero pitch rate and fixed $\alpha_\td$. Let $\Ntd(\X)$ be the touchdown force \eqref{eq:Ntd}.
\begin{enumerate}
\item[(i)] The stance and flight equations are reversible with respect to $\mathcal{R}:(x,z,\dot x,\dot z)\mapsto(-x,z,\dot x,-\dot z)$ combined with time reversal.
\item[(ii)] If $\Ntd=0$, $N_{\mathrm{des}}>0$ on $(0,\pi/(2\omega_0)]$ and the stance trajectory satisfies $\thv(\pi/(2\omega_0))=0$, the stride is symmetric. Liftoff then occurs at $t=\pi/\omega_0$ with $\rv=\rvz$ and without force step, and the apex state is a fixed point.
\item[(iii)] Let $\tilde P$ denote the apex map in which liftoff is replaced by the event $\rv=\rvz$ at $t=\pi/\omega_0$. The fixed points of $\tilde P$ that satisfy $\thv(\pi/(2\omega_0))=0$ form one-parameter families. Along such a family, $P(\X)-\X=O(\Ntd^2)$, and the stride energy changes by
\begin{align}
\Delta E&=\frac{\Ntd^2}{2\kappa}+O(\Ntd^3),\label{eq:drift}\\
\kappa&=-\frac{\dot N_{\mathrm{des}}}{\dot r^{v}}\Big|_{t=\pi/\omega_0}\nonumber\\
&=k^v-3m(\dot\theta^{v})^2+\frac{3mg\,\dot\theta^{v}\sin\thv}{\dot r^{v}}.\label{eq:kappa}
\end{align}
\item[(iv)] At a fixed point of type (ii), the Jacobians of $P$ and $\tilde P$ coincide. Hence $\partial P/\partial(z_a,\dot x_a)$ has the eigenvalue one.
\end{enumerate}
\end{proposition}

\begin{proof}
(i) The force $N_{\mathrm{des}}$ in \eqref{eq:template} depends on $\thv$ through $\cos\thv$ and on $\dot\theta^{v}$ through its square, and $\mathcal{R}$ maps $(\thv,\dot r^{v},\dot\theta^{v})$ to $(-\thv,-\dot r^{v},\dot\theta^{v})$ under time reversal. Ballistic flight is reversible as well. (ii) By \eqref{eq:rv}, $\dot r^{v}=0$ at $t=\pi/(2\omega_0)$. With $\thv=0$ there, the state lies on the fixed set of $\mathcal{R}$, so the trajectory is symmetric about this instant. At the mirrored time $t=\pi/\omega_0$, the state is $\mathcal{R}$ applied to the touchdown state. In particular, $\rv=\rvz$ and $N_{\mathrm{des}}=\Ntd=0$, so liftoff occurs there and the liftoff velocity mirrors the touchdown velocity. The next apex therefore equals the current one. (iii) The stance flow is reversible by (i), and $\rv(\pi/\omega_0)=\rvz$ for every stride by \eqref{eq:rv}. A stride that passes through the symmetry set at $t=\pi/(2\omega_0)$ therefore reaches the mirror image of its touchdown state at $t=\pi/\omega_0$, and it is a fixed point of $\tilde P$. This is one scalar condition, so these fixed points form curves, as in Proposition~\ref{prop:symmetry}. On such a curve, symmetry gives $N_{\mathrm{des}}(\pi/\omega_0)=\Ntd$. For $\Ntd>0$, stance continues after $\pi/\omega_0$ until $N_{\mathrm{des}}=0$, during an interval $\Delta t=\Ntd/(\kappa\dot r^{v})+O(\Ntd^2)$, with $\dot r^{v}$ evaluated at $\pi/\omega_0$. During this interval, the true and the $\tilde P$ trajectories differ by an acceleration of order $\Ntd/m$, so their states after liftoff differ by $O(\Ntd^2)$. Since $\tilde P(\X)=\X$ on the curve, $P(\X)-\X=O(\Ntd^2)$. The extra work done by the GRF during $\Delta t$ is $\int N_{\mathrm{des}}\dot r^{v}dt=\Ntd^2/(2\kappa)+O(\Ntd^3)$, which gives \eqref{eq:drift}. For $\Ntd<0$, the stride is inadmissible; formally, the same argument applies to an early liftoff and gives the same result. (iv) For a general stride near the fixed point, the same argument with $N_\lo=N_{\mathrm{des}}(\pi/\omega_0)$ in place of $\Ntd$ gives $P-\tilde P=O(N_\lo^2)$. Since $N_\lo$ is smooth and vanishes at the fixed point, $P-\tilde P=O(\|\X-\X^*\|^2)$, so the first derivatives of $P$ and $\tilde P$ coincide there. Differentiating $\tilde P(\X(s))=\X(s)$ along the family gives the eigenvalue one. The expression for $\kappa$ follows from differentiating $N_{\mathrm{des}}$ in \eqref{eq:template} and using \eqref{eq:embedded}.
\end{proof}

Proposition~\ref{prop:natural} explains the structure that we observed numerically. It does not exclude non-symmetric gaits. An unconstrained search supports this picture. We started Newton iterations on $P(\X)=\X$ from $10\times10$ initial apex states ($z_a\in[1.0,2.4]$\,m, $\dot x_a\in[2,7]$\,m/s) for twelve combinations of $k^v\in\{12,20,30\}$\,kN/m, $\theta^*\in\{0^\circ,10^\circ\}$ and $\alpha_\td$. All 843 converged runs ended at the natural gait, within $5\times10^{-4}$ in $(z_a,\dot x_a)$. Because the fixed point is degenerate, a residual of $10^{-8}$ corresponds to a position error of order $10^{-4}$. A natural gait is isolated for fixed $\alpha_\td$ and $\theta^*$, because it requires both the symmetry condition and $\Ntd=0$. It is nevertheless non-hyperbolic, because it lies on the family of the reversible map $\tilde P$, from which $P$ deviates only to second order. Along this family, the residual $P(\X)-\X$ grows quadratically, and \eqref{eq:drift} shows that it has a fixed sign in energy. Members of the family with $\Ntd>0$ gain energy in every stride, and those with $\Ntd<0$ are inadmissible. The natural gait therefore lies at the boundary of the admissible part of its family. These gaits need no force step at touchdown, so the main physical drawback of the linear template disappears at steady state.

The proposition also reduces the computation of natural gaits to a scalar equation. The condition $\Ntd=0$ fixes the angular rate at touchdown,
\begin{equation}
\dot\theta^{v}_\td=-\sqrt{g\cos\thv_\td/\rvz}.
\label{eq:dthv_nat}
\end{equation}
The symmetry condition $\thv(\pi/(2\omega_0))=0$ then determines $\dot r^{v}_\td$. With the exact sweep \eqref{eq:sweep}, it becomes a closed-form equation in $a=\dot r^{v}_\td/\omega_0$.

\begin{corollary}[Closed-form natural gaits]\label{cor:gait}
Approximate the gravitational moment in the first half of stance as decreasing linearly to zero at mid-stance, and use its mean momentum in the sweep. Then the natural gait for given $(\alpha_\td,\theta^*,k^v)$ approximately satisfies
\begin{equation}
\begin{gathered}
\thv_\td+\frac{\hat h}{\omega_0}\,\mathcal{J}\Big(\frac{\pi}{2};\rvz,a\Big)=0,\\
\hat h=(\rvz)^2\dot\theta^{v}_\td+\frac{\pi g\rvz\sin\thv_\td}{6\omega_0},
\end{gathered}
\label{eq:cf_gait}
\end{equation}
with $\rvz$, $\thv_\td$ from \eqref{eq:td_geometry}, $\dot\theta^{v}_\td$ from \eqref{eq:dthv_nat} and $\mathcal{J}(\pi/2;\rho,a)=(\rho\,\mathcal{I}(\pi/2)-a/\rho)/(\rho^2-a^2)$. Here $\hat h$ is the mean momentum over the first half of stance, and, as in GC-L, it replaces $h(t)$ in the sweep. The apex state follows from the touchdown velocity by ballistic flight.
\end{corollary}

\begin{figure*}[!t]
\centering
\includegraphics[width=\textwidth]{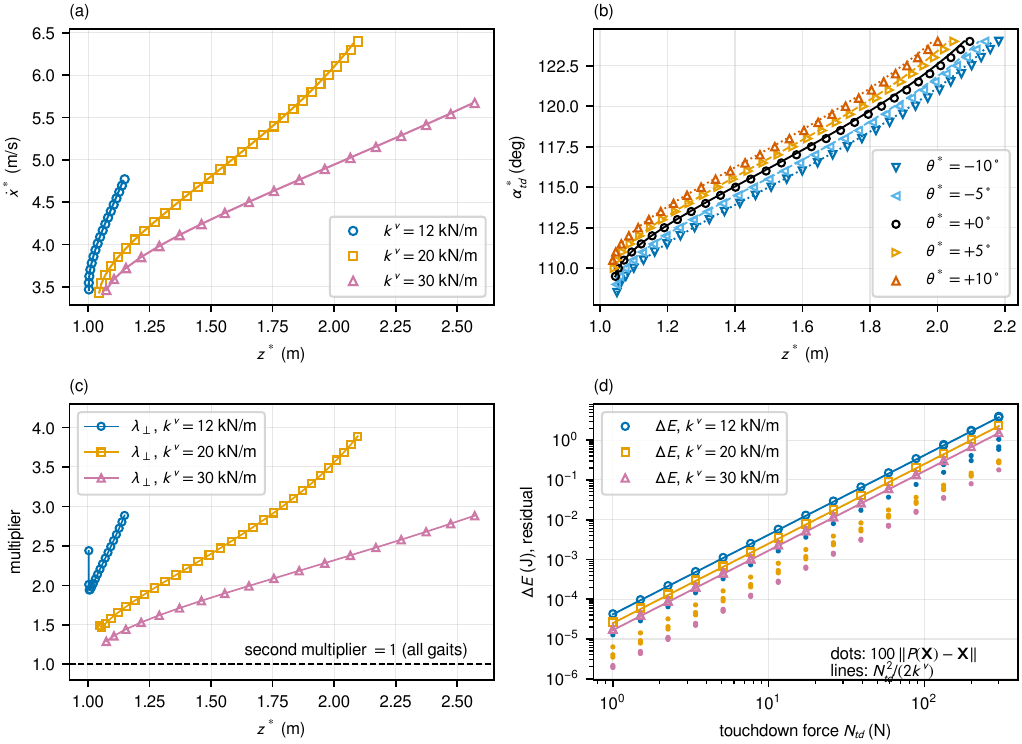}
\caption{Natural gaits of the passive embedded runner. (a) Apex speed versus apex height for three virtual stiffnesses at upright posture. (b) Touchdown angle versus apex height for five postures ($k^v=20$\,kN/m). (c) Transverse multiplier; the second multiplier equals one at all gaits. Markers in (a)--(c): numerical gaits; lines in (a) and (b): closed form \eqref{eq:cf_gait}; lines in (c) join the numerical values. (d) Energy change per stride (hollow markers) and $100\,\|P(\X)-\X\|$ (dots) along the symmetric families of $\tilde P$ at $\alpha_\td\in\{110^\circ,113^\circ,116^\circ\}$, versus the touchdown force $\Ntd$; lines show $\Ntd^2/(2k^v)$. The residual combines components in m and m/s}
\label{fig:fp_embedded}
\end{figure*}

We computed the natural gaits numerically for $k^v\in\{12,20,30\}$\,kN/m, $\theta^*\in\{0^\circ,\pm5^\circ,\pm10^\circ\}$ and $\alpha_\td\in[100^\circ,124^\circ]$ in steps of $0.5^\circ$. For each case, we fixed $\dot\theta^{v}_\td$ by \eqref{eq:dthv_nat} and solved the symmetry condition for $\dot r^{v}_\td$ with Brent's method. This gave 330 gaits. All of them are fixed points of the full model with residuals below $4\times10^{-14}$. At liftoff, $\rv-\rvz$ is below $2\times10^{-15}$\,m. One multiplier of the $(z_a,\dot x_a)$ block equals one within $2\times10^{-8}$. The other one, $\lambda_\perp$, ranges from 1.29 to 4.09, so all natural gaits are unstable. Figure~\ref{fig:fp_embedded} summarizes the results.

The closed form \eqref{eq:cf_gait} predicts the apex height within 0.82\% and the apex speed within 0.41\% for all 330 gaits, up to apex heights of 2.6\,m. Without the gravity term in $\hat h$, the errors grow to 9.0\% in height and 6.2\% in speed, and \eqref{eq:cf_gait} has no root for 20 of the 330 cases. Softer virtual legs give faster gaits at a given height [Fig.~\ref{fig:fp_embedded}(a)]. Posture hardly changes the gait speed: at $z^*=1.2$\,m, the speed varies by 0.03\,m/s between $\theta^*=-10^\circ$ and $10^\circ$. Instead, the required touchdown angle increases by about $0.57^\circ$ per $5^\circ$ of forward lean [Fig.~\ref{fig:fp_embedded}(b)], because lean changes the touchdown geometry \eqref{eq:td_geometry}. Except at the lowest heights, the transverse multiplier grows with apex height, and it decreases with stiffness [Fig.~\ref{fig:fp_embedded}(c)]. Figure~\ref{fig:fp_embedded}(d) confirms \eqref{eq:drift}. Along the symmetric families, both the energy gain and the residual grow as $\Ntd^2$. The ratio of the energy gain to $\Ntd^2/(2k^v)$ lies between 1.00 and 1.08 for $\Ntd\in[1,300]$\,N. Evaluating $\kappa$ with \eqref{eq:kappa} reproduces this ratio within 0.1\%.

%% ===================================================================
\section{Stride-level control with the analytic map}\label{sec:mapcontrol}
%% ===================================================================

The natural gaits are unstable, and the analytic map of Section~\ref{sec:approx} is fast and accurate. We therefore use it for deadbeat stride control.

The touchdown angle redirects the COM velocity. A second input must change the stride energy. The virtual stiffness $k^v$ is a poor choice. Near natural gaits, $\alpha_\td$ and $k^v$ move the apex state in almost the same direction. At the natural gait with $z^*=1.23$\,m and $k^v=20$\,kN/m, the two columns of the input matrix are parallel to within the accuracy of finite differences, and their scaled condition number (scales $5^\circ$ and 25\%) exceeds $10^4$. We therefore add a thrust $\delta$ to the template: during decompression ($\dot r^{v}>0$), the rest length becomes $\rvz+\delta$, similar to the thrust of Raibert's hoppers \cite{raibert1986legged}. The radial motion stays exact. After the bottom at $t_b=\pi/(2\omega_0)$,
\begin{equation}
\rv(t)=\rvz+\delta+(a-\delta)\cos\omega_0(t-t_b),
\end{equation}
and the sweep integral splits into two terms of the form \eqref{eq:J}. The liftoff estimate is evaluated at $\rv=\rvz+\delta$. A thrust of 3\,cm changes the stride energy by about 130\,J, and the input matrix of $(\alpha_\td,\delta)$, scaled by $5^\circ$ and 3\,cm, has a condition number of about 5. On 430--600 random valid strides per value of $\delta\in[-3,5]$\,cm, without the admissibility filter, the mean errors of the GC-L map stay within 0.6--0.7\% (liftoff position), 1.1--1.2\% (liftoff velocity) and 0.7--0.9\% (apex position).

\subsection{Deadbeat control with error feedback}

Let $\hat P_s(\X,u)$ denote the $(z_a,\dot x_a)$ part of the analytic map with $\mathbf{u}=[\alpha_\td,\delta]^T$. At apex $k$, the controller solves
\begin{equation}
\hat P_s(\X_k,\mathbf{u}_k)=\X_{s,k}^{\mathrm{ref}}-\mathbf{b}_k
\label{eq:deadbeat}
\end{equation}
for $\mathbf{u}_k$ by Newton iterations with $\alpha_\td\in[95^\circ,130^\circ]$ and $\delta\in[-6,10]$\,cm. The term $\mathbf{b}_k$ is zero for the plain analytic controller. With error feedback, $\mathbf{b}_k$ is a filtered estimate of the map error,
\begin{equation}
\begin{gathered}
\mathbf{b}_{k+1}=\mathbf{b}_k+\beta\big(\X_{s,k+1}-\hat P_s(\X_k,\mathbf{u}_k)-\mathbf{b}_k\big),\\ \mathbf{b}_0=\mathbf{0},
\end{gathered}
\label{eq:bias}
\end{equation}
with $\beta=0.5$. If the map error varies slowly, this feedback removes the steady-state error, like an integral action or a disturbance observer. A gain $\beta=1$ removes the steady-state error faster, but in the scenario below it overshoots more after the pitch-rate kick (0.21 instead of 0.16\,m/s). As an ideal reference, we also solve \eqref{eq:deadbeat} with the numerical return map and $\mathbf{b}_k=0$. In all cases, $k^v=20$\,kN/m, the stance PD law \eqref{eq:pd} holds the pitch reference $\theta^*$, and the hip torque is limited as in Section~\ref{sec:limits}.

\subsection{Results}

\begin{figure*}[!t]
\centering
\includegraphics[width=\textwidth]{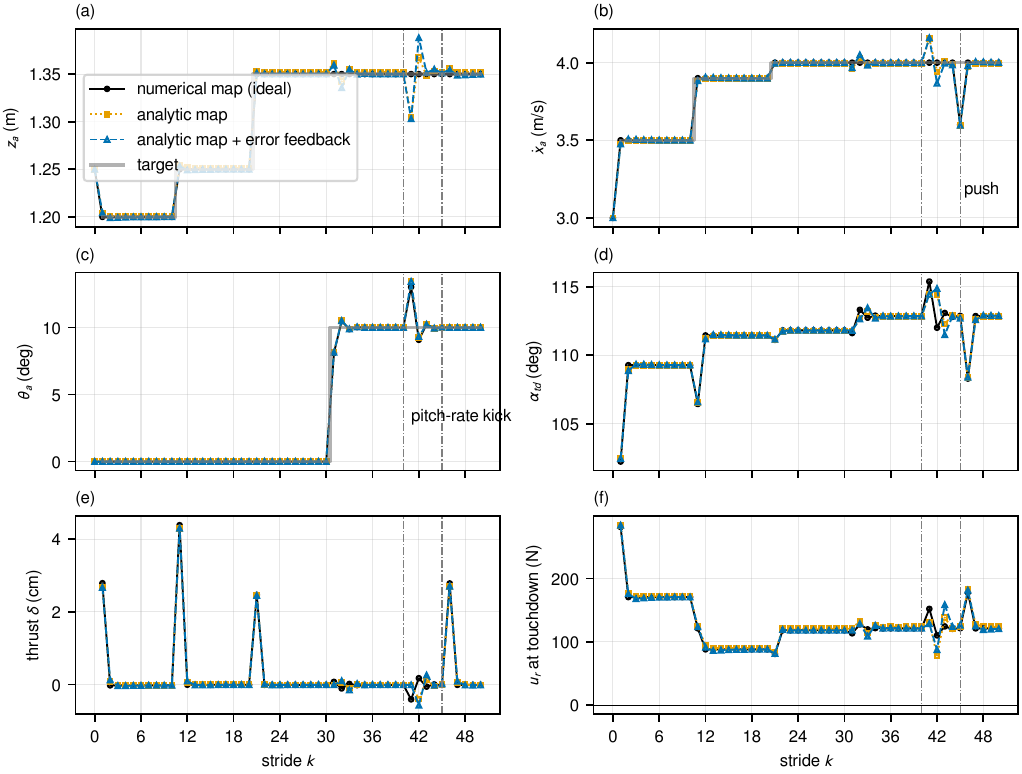}
\caption{Deadbeat stride control of the embedded runner ($k^v=20$\,kN/m). (a)--(c) Apex height, speed and pitch with targets (gray); (d) touchdown angle; (e) thrust; (f) actuator force $u_r$ at touchdown. Vertical dash-dotted lines: pitch-rate kick (stride 40) and push (stride 45)}
\label{fig:ctrl_scenario}
\end{figure*}

Figure~\ref{fig:ctrl_scenario} shows a 50-stride scenario. The targets change height and speed at strides 1, 11 and 21. At stride 31, the pitch reference changes from $0^\circ$ to $10^\circ$. A pitch-rate kick of $60^\circ$/s and a push of $-0.4$\,m/s follow at strides 40 and 45. All targets lie below the natural-gait speed for their height, so the steady strides are admissible. The numerical-map controller reaches every target in one stride, also after the lean change and the disturbances. In the first stride, the analytic-map controller misses the speed target by 0.024\,m/s, which is 4.7\% of the commanded change. Without error feedback, it keeps a steady-state error of up to 2\,mm in height and 5\,mm/s in speed. With error feedback, this error falls below $10^{-4}$ (in m and m/s) within ten strides. The lean change causes speed errors of up to 0.05\,m/s, because the PD torque that rotates the trunk violates the CPP condition. For the same reason, the pitch-rate kick causes an error of 0.16\,m/s in the next stride. The analytic controllers recover from both within four strides, and from the push within two. The actuator force at touchdown stays between 78 and 284\,N. Thrust is used mainly at speed changes.

\begin{figure}[!tb]
\centering
\includegraphics[width=0.92\columnwidth]{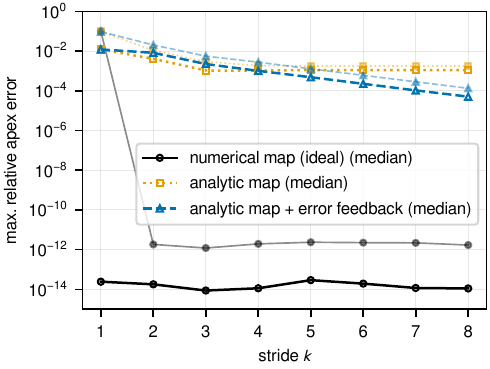}
\caption{Largest relative apex error in 400 random trials: median (bold lines) and 95th percentile (faint lines)}
\label{fig:ctrl_mc}
\end{figure}

To quantify these results, we ran 400 trials with random initial apex states ($z_a\in[1.1,1.4]$\,m, $\dot x_a$ between 2.5\,m/s and the natural speed, $|\theta_a|\le10^\circ$, $|\dot\theta_a|\le30^\circ$/s), pitch reference $\theta^*=0$, and random admissible targets. Figure~\ref{fig:ctrl_mc} shows the largest relative error of $(z_a,\dot x_a)$. No trial failed. After the first stride, the median errors are 1.2\% for the analytic controllers and below $10^{-13}$ for the numerical one. The 95th percentiles are about 10\% for all controllers, because some targets cannot be reached in one stride within the input bounds. Without error feedback, the analytic controller settles at a median error of 0.11\% (95th percentile 0.18\%). With error feedback, the median error after eight strides is $5\times10^{-5}$ (95th percentile $1.3\times10^{-4}$) and still decreasing. Only 0.5\% of the first strides and 0.1\% of the later strides had a negative touchdown force.

%% ===================================================================
\section{Discussion}\label{sec:discussion}
%% ===================================================================

CPP is the VPP law with the pivot at the COM. A VPP above the COM turns the trunk into a pendulum and stabilizes posture passively in stance \cite{maus2010upright,sharbafi2013robust,sun2023posture,karagoz2022vpp}. Human running, however, shows an intersection point below the COM \cite{drama2020postural,firouzi2022human}, and its height trades hip work against leg work \cite{drama2020trunk}. CPP sits between these cases. It gives up passive posture stability and gains an exact decoupling, which makes the COM problem a central-force problem. Posture is then stabilized by feedback, which is consistent with the finding that a GRF intersection point is not necessary for stability \cite{schreff2023vpp}. With bounded torque and friction, this feedback recovers from large pitch disturbances (Section~\ref{sec:control}). The ASLIP controller of Poulakakis and Grizzle also uses both actuators to impose SLIP on the COM while regulating posture \cite{poulakakis2009spring}. With CPP, the same goal follows from a static torque law and a single radial force law, and the embedded dynamics do not depend on posture at all. Posture becomes an additional gait parameter.

The choice of template involves a trade-off. The SLIP template keeps the GRF continuous and is always admissible. Its stance map is not integrable, but established approximations apply directly. The linear template makes the radius and the constant-momentum sweep exact, and the GC-L map is the most accurate one we tested. Its price is the touchdown force step and the admissibility condition $\Ntd\ge0$, which limits the speed for a given geometry and stiffness. Proposition~\ref{prop:natural} shows that the step vanishes at natural gaits, so it only appears in transients. In the control experiments, the actuator force at touchdown stayed below $0.4\,mg$. Softer virtual legs raise the admissible speed. Other radial laws could combine the advantages of both templates, as long as the radial motion stays integrable.

Piovan and Byl \cite{Piovan2016} and Hamza\c{c}ebi and Morg\"ul \cite{Hamzacebi2017} used PFL to simplify the radial dynamics of actuated SLIP models; the latter obtained exact solutions with a leg and a hip actuator. Both studies consider a point mass on a leg, without a trunk, posture or pitch dynamics. Our setting is considerably more general. The model has a rigid trunk with a hip offset, the posture can take any value during running, and the hip torque must balance the trunk while the leg shapes the COM motion. CPP is the step that makes PFL possible in this setting. Without it, the GRF has a component normal to the virtual leg, and the angular dynamics are not those of a point mass. Beyond the embedding, we also give the structure of the return map, gait families from reversibility, approximate analytic maps for asymmetric strides and arbitrary posture, the natural gaits of the embedded runner, and map-based control. Selvitella and Foster \cite{selvitella2024approximate} reach an elementary radial motion by assuming a constant angular velocity, and they treat symmetric gaits. In our case, the radial motion is exact by construction, the constant-momentum sweep is integrated exactly, and GC-L handles asymmetric strides. Perturbation methods \cite{Yu2012,yu2019seeking,chen2023articulated,kilic2023novel} and learned stance maps \cite{ozturk2026physics} target the passive SLIP and its extensions. They could be applied to the SLIP template, which CPP embeds exactly for any posture.

Both runners studied here have non-hyperbolic periodic gaits. In energetically conservative legged models, gait families follow from energy conservation \cite{raff2022connecting}, and symmetries organize the possible gaits \cite{Altendorfer2004,ding2024breaking}. The CPP runner is not conservative, yet reversibility at upright posture gives one-parameter families. For the embedded runner, the families belong to the reversible map $\tilde P$, and the actual gaits are isolated points on them. In both cases, a stride-level controller with two inputs removes the unit multiplier. Near such gaits, small errors of an approximate map can shift its fixed points considerably along the family. Map-based controllers should therefore use feedback, such as the error feedback in Section~\ref{sec:mapcontrol}, rather than rely on the predicted gait.

The GC-L map takes a few microseconds per stride in our implementation, more than 300 times less than the fixed-step simulation and several orders of magnitude less than adaptive integration. It depends smoothly on $(\alpha_\td,\theta_\td,k^v,\delta)$. It can therefore support stride-level deadbeat or model-predictive control \cite{Wu2013,arslan2012reactive}, gait optimization, and reduced-order control of humanoid robots.

The approach also opens several directions. CPP needs only a static hip-torque law and a radial leg force, so it can be implemented on robots with a series or parallel elastic leg and a torque-controlled hip. In biomechanics, it provides a reference case between the VPP above the COM and the intersection point below the COM observed in human running \cite{drama2020postural,firouzi2022human}; deviations from CPP can be studied as perturbations of an analytically tractable runner. The frontal-plane VPP above the COM reported in \cite{firouzi2022human} suggests a three-dimensional extension, in which the same decoupling would separate trunk orientation from COM motion in space. Finally, because CPP embeds SLIP exactly at any posture, all analytic and learned SLIP maps \cite{Saranli2010,selvitella2024approximate,ozturk2026physics} transfer directly to a model with a trunk.

The study has several limitations. The leg is massless, so the trunk cannot be controlled in flight, and leg placement is instantaneous. The torque and friction limits of Section~\ref{sec:limits} only saturate the hip torque; the actuator force $u_r$ is not bounded, and the toe does not slip. The embedding requires an actuator that can produce force steps and both inject and remove energy. Negative damping in the stride-level LQR also requires actuation. The pitch gains were tuned at one gait. Finally, all results are numerical simulations of a human-sized model, and experiments remain to be done.

%% ===================================================================
\section{Conclusion}\label{sec:conclusion}
%% ===================================================================

We studied the central pivot point policy for a body-attached spring-mass runner. CPP decouples pitch from the COM motion exactly and turns stance into a central-force problem. The resulting return map has structural unit multipliers, and upright gaits form symmetric one-parameter families. A stance PD law combined with a stride-level LQR stabilizes all computed gaits, also with bounded hip torque and friction. Under CPP, any radial force law in virtual-leg coordinates can be embedded independently of posture. For a linear radial template, the exact radial solution allows an exact sweep integral. With a linear-torque gravity correction, the resulting analytic map predicts liftoff and apex states with mean errors between 0.3\% and 1.6\% on admissible strides, including asymmetric strides, leaning postures and nonzero pitch rates. The natural gaits we computed are symmetric, need no force step at touchdown, and are non-hyperbolic because the return map is reversible up to second order in the touchdown force. They follow from a scalar equation. The analytic map supports deadbeat stride control, and a filtered error feedback removes its steady-state error. Natural next steps are templates that avoid the force step while keeping the radial motion integrable, leg dynamics, and experiments on a robot with a trunk.

\begin{acknowledgments}
The authors thank Prof.\ Ulu\c{c} Saranl{\i} for helpful discussions in the early stages of this work. This work was supported by Middle East Technical University under project ADEP-301-2025-11613. The ideas, derivations, simulations and analyses at the core of this work were developed in the MSc thesis of the first author \cite{sever2022thesis}. In preparing this manuscript from the thesis, the authors used a large language model (Claude, Anthropic) to help re-implement some of the simulations, repeat and extend some of the analyses, and edit the text. The authors directed this work, checked all results and take full responsibility for the content.
\end{acknowledgments}

\section*{Author Declarations}

\subsection*{Conflict of Interest}
The authors have no conflicts to disclose.

\subsection*{Author Contributions}
\textbf{{\.I}zel Sever:} Conceptualization (equal); Methodology (equal); Software (equal); Investigation (lead); Formal analysis (equal); Writing -- original draft (lead); Writing -- review \& editing (equal).
\textbf{O. Kaan Karag\"oz:} Conceptualization (equal); Methodology (equal); Software (equal); Formal analysis (equal); Writing -- review \& editing (equal).
\textbf{M. Mert Ankaral{\i}:} Conceptualization (equal); Methodology (equal); Formal analysis (equal); Supervision (lead); Funding acquisition (lead); Writing -- review \& editing (lead).

\section*{Data Availability}
The data and code that support the findings of this study will be made openly available prior to publication.

\clearpage
\appendix

\section{Equations of motion}\label{app:eom}

The Lagrangian of the trunk is $L=\tfrac12 m(\dot x^2+\dot z^2)+\tfrac12 J\dot\theta^2-mgz$. The leg force and hip torque enter through the virtual work $F_r\,\delta r+\tau\,\delta\psi$, which gives \eqref{eq:stance} with
\begin{align}
\nabla_{\mathbf{q}} r&=\frac{1}{r}\begin{bmatrix}x_h\\ z_h\\ \dhip(z\sin\theta-x\cos\theta)\end{bmatrix},\\
\nabla_{\mathbf{q}}\psi&=\frac{1}{r^2}\begin{bmatrix}-z_h\\ x_h\\ x^2+z^2-\dhip(x\sin\theta+z\cos\theta)\end{bmatrix}.
\end{align}
The first two components give the GRF \eqref{eq:grf}.

\section{Geometry of the CPP torque}\label{app:geometry}

Consider the triangle formed by the toe, the hip and the COM [Fig.~\ref{fig:virtualleg}]. The leg vector from the toe to the hip is $r[\cos\alpha,\sin\alpha]^T$, and the hip-to-COM vector is $\dhip[\sin\theta,\cos\theta]^T$. Projecting the virtual leg on the leg direction and on its normal gives
\begin{equation}
\rv\cos\gamma=r-\dhip\cos\psi,\qquad \rv\sin\gamma=\dhip\sin\psi,
\end{equation}
so that $\tan\gamma=\dhip\sin\psi/(r-\dhip\cos\psi)$, which is \eqref{eq:tau_cpp}. Projecting the leg on the virtual-leg direction gives $r\cos\gamma=\rv-\dhip\cos\Phi$, which is \eqref{eq:triangle}. The virtual touchdown pair \eqref{eq:td_geometry} follows from $x_\td=r_0\cos\alpha_\td+\dhip\sin\theta_\td$ and $z_\td=r_0\sin\alpha_\td+\dhip\cos\theta_\td$.

\section{Coordinate changes}\label{app:transform}

With the toe at the origin, $T_{c\to v}$ maps the Cartesian COM state to
\begin{align}
\rv&=\sqrt{x^2+z^2}, & \dot r^{v}&=\frac{x\dot x+z\dot z}{\rv},\\
\thv&=\operatorname{atan2}(-x,z), & \dot\theta^{v}&=\frac{x\dot z-z\dot x}{(\rv)^2},
\end{align}
and $T_{v\to c}$ is its inverse,
\begin{align}
x&=-\rv\sin\thv, & \dot x&=-\dot r^{v}\sin\thv-\rv\dot\theta^{v}\cos\thv,\\
z&=\rv\cos\thv, & \dot z&=\dot r^{v}\cos\thv-\rv\dot\theta^{v}\sin\thv.
\end{align}
The pitch states are common to both coordinate sets.

\section{Baseline angular approximations}\label{app:angular}

The baseline maps NC, GC-M and GC-T use the radial solution \eqref{eq:rv} and expand $1/(\rv)^2$ to first order about $\rvz$, $1/(\rv)^2\approx(\rvz)^{-2}\big(1-2(\rv-\rvz)/\rvz\big)$. With a constant momentum $p=m(\rvz)^2\omega$, this gives
\begin{align}
\dot\theta^{v}(t)&=\omega\Big(1-\frac{2\dot r^{v}_\td}{\omega_0\rvz}\sin\omega_0 t\Big),\\
\thv(t)&=\thv_\td+\omega t+\frac{2\omega\dot r^{v}_\td}{\omega_0^2\rvz}\big(\cos\omega_0 t-1\big).
\end{align}
The liftoff time is \eqref{eq:tlo} with $\thv(\pi/\omega_0)$ from this expansion. NC uses $\omega=\dot\theta^{v}_\td$. The corrections use the uncorrected liftoff estimate $\rv_\lo$, $\thv_\lo$:
\begin{align}
\text{GC-M:}\quad\hat p&=p_\td+\frac{mg\,t_s}{4}\big(\rvz\sin\thv_\td+\rv_\lo\sin\thv_\lo\big),\\
\text{GC-T:}\quad\hat p&=p_\td+\frac{mg\,t_s\,\bar r^{v}}{2}\big(\sin\thv_\td+\sin\thv_\lo\big),\\
\bar r^{v}&=\rvz+\frac{\dot r^{v}_\td(1-\cos\omega_0 t_s)}{\omega_0^2t_s},
\end{align}
and $\hat\omega=\hat p/(m(\rvz)^2)$ replaces $\omega$ in both the angle and the rate. The liftoff time is not updated after the correction.

\bibliography{references}

\end{document}